\pdfoutput=1
\documentclass[11pt]{article}

\usepackage[final]{EMNLP2023}

\usepackage{times}
\usepackage{latexsym}

\usepackage[T1]{fontenc}
\usepackage[utf8]{inputenc}

\usepackage{microtype}

\usepackage{inconsolata}

\usepackage{graphicx}
\usepackage{amsmath}
\usepackage{xurl}
\usepackage{amssymb}
\usepackage{amsthm}
\usepackage{mathtools}
\usepackage{todonotes}
\usepackage{color}

\usepackage{caption}
\usepackage{subcaption}
\usepackage{dsfont}

\newtheorem{definition}{Definition}
\newcommand{\Sr}{\mathcal{S}_r}
\newcommand{\ptrain}{p^{\text{train}}}
\newcommand{\ptrue}{p^{\text{true}}}
\newcommand{\ptruedef}{p^{\emph{true}}}
\newcommand{\pagguni}{p^{\text{agg\_uni}}}
\newcommand{\paggmodel}{p^{\text{agg\_model}}}

\title{Geographical Erasure in Language Generation}

\author{Pola Schw\"obel \\
  Amazon\\
  Berlin, Germany \\
  \texttt{schwobel@amazon.de} \\\And
  Jacek Golebiowski \\
  Amazon \\
  Berlin, Germany \\
   \texttt{jacekgo@amazon.de} \\\And
  Michele Donini \\
  Amazon \\
  Berlin, Germany \\
   \texttt{donini@amazon.de} \\
   \AND
    C\'edric Archambeau\thanks{~~Work done while at Amazon.} \\
  Helsing \\
  Berlin, Germany \\
  \texttt{cedric.archambeau@helsing.ai} \\\And
  Danish Pruthi$^*$  \\
  Indian Institute of Science (IISc) \\
  Bangalore, India \\
   \texttt{danishp@iisc.ac.in} \\
  }

\begin{document}
\maketitle
\begin{abstract}

Large language models (LLMs) encode vast amounts of world knowledge. However, since these models are trained on large swaths of internet data, they are at risk of inordinately capturing information about dominant groups. This imbalance can propagate into generated language. In this work, we study and operationalise a form of \emph{geographical erasure}, wherein language models underpredict certain countries. 
We demonstrate consistent instances of erasure across a range of LLMs.
We discover that erasure strongly correlates with low frequencies of country mentions in the training corpus. 
Lastly, we 
mitigate erasure by finetuning using a custom objective.\footnote{Code available at \url{https://github.com/amazon-science/geographical-erasure-in-language-generation}.} 
\end{abstract}

\section{Introduction}\label{sec:intro}

Large pretrained models
serve as
base models for 
many downstream NLP applications, including question-answering,
dialogue, common-sense reasoning, classification, tagging, translation, summarisation, and generation~\cite{devlin2018bert, brown2020language, chowdhery2022palm}. 
Despite
their 
increasing
utility,
there are 
concerns 
about 
how they reflect 
and amplify
biases in the training data. 
For instance, unfiltered data originating 
from the internet is known 
to be rife
with toxic,
misogynistic,
and stereotyping content.
Many studies
highlight
biases 
in model outputs, 
primarily concerning 
\emph{representational harms} \cite{barocas2017problem},
where 
a section of society 
(e.g., women, LGBTQ+ communities)
are represented in poor light, 
or are
ignored by the system~\cite{bolukbasi2016man, guo2021detecting, may2019measuring, tan2019assessing}.
While important, such studies predominantly 
examine biases 
related to race, gender, occupation and sexual orientation.

\begin{figure}[t]
\includegraphics[width=0.8\linewidth]{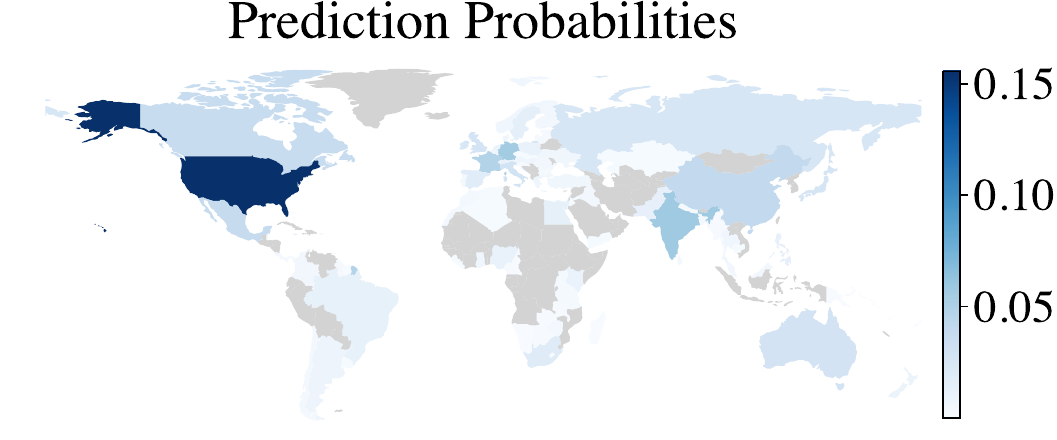}
\includegraphics[width=0.8\linewidth]{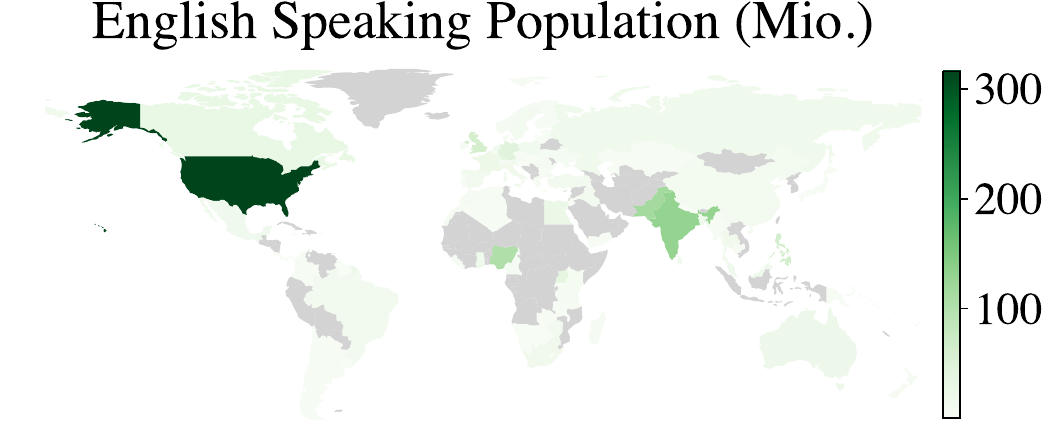}
\includegraphics[width=0.8\linewidth]{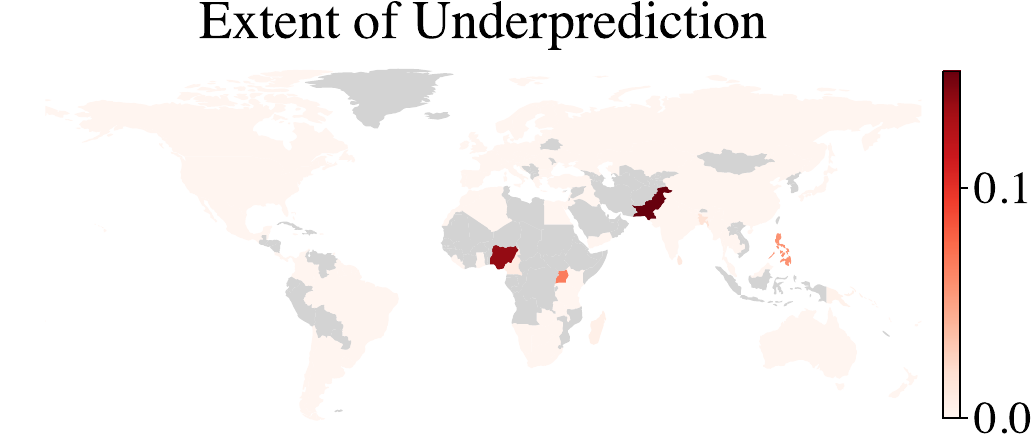}
\caption{\textbf{Some countries are vastly underpredicted compared to their English speaking populations.} \textit{Top:} Country probabilities assigned by GPT-NeoX when prompted with ``I live in''. \textit{Middle:} English speaking populations per country. \textit{Bottom:} Countries experiencing erasure, i.e.\, underprediction compared to their population by at least a factor $3$ (see \S\ref{sec:method}). Data is missing for grey countries (see \S\ref{sec:limitations}).}\label{fig:teaser}
\end{figure}

An important---and often overlooked---aspect 
of inclusive
model development
is \emph{geographical} inclusion. 
This is particularly important at a time  
when 
most 
large-scale
model training 
efforts 
come from a small set 
of regions. 
Further, these models
are 
trained using internet data,
whose access, in the first place, is unequally distributed~\cite{blank2017digital, pew_research_internet_access}. 
Minimising cultural and geographic identities is referred to as erasure \citep{roche2019articulating} and is studied by linguists and social scientists in the context of imperialism and colonialism, where “people are silenced in the historical record [...], their contemporary presence rendered invisible, and their existence written out of the future” \citep{roche2019articulating}.
Automated systems and their developers
exclude 
certain groups unintentionally,
but the risk of being ``written out of the future'' remains pressing: the produced content is fed back into the internet. 
In lifelong learning setups, 
the generated content becomes the training data of tomorrow's models, closing the vicious circle of 
reinforcing social hierarchies (see \S4.5 of \citep{sheng2021societal}). 

In this paper,
we reveal
instances 
of \emph{geographical erasure},
wherein 
models
underpredict
geographical regions (see Fig.~\ref{fig:teaser}).
For instance, GPT2 assigns nine
times higher likelihood to “I live in \underline{Canada}” than “I live in \underline{Pakistan}”, whereas Pakistan's English-speaking population is almost four times to that of
Canada.
By comparing model outputs
with population statistics,
we 
operationalise 
geographical erasure (\S\ref{sec:method}).
Using this measure,
we first demonstrate the 
existence of 
erasure for several countries across different prompt formulations (\S\ref{sec:exp1}). Studying the consistency across a range of language models from the GPT and LLaMA model families, we find that several countries --- Nigeria, Pakistan, Eswatini, Uganda and Madagascar --- are affected by erasure under \textit{all} models (\S\ref{sec:erasure_who}). 
Following related work \citep{lin2021truthfulqa, rae2021scaling, nadeem2020stereoset}, 
we study the impact of model size on the extent of erasure, and find it to be small --- erasure occurs across all sizes (\S\ref{sec:exp2}).
To identify the 
causes of 
erasure,
we 
compute the unigram
frequencies of countries
in the training corpus (\S\ref{sec:exp3}). They closely 
match our model predictions indicating that the composition of training data is a main source of erasure. Lastly, we alleviate 
erasure
via supervised finetuning.
We study the impact of mitigation
on generation quality as measured in perplexity on Wikitext-2-v1. Our finetuning strategy proves to be an effective mitigation mechanism which generalises and has small impact on generation quality  (\S\ref{sec:exp4}).

\section{Related Work}\label{sec:related_work}
% general overview, allocational vs. representational harms
The literature on fairness in machine learning distinguishes between representational and allocational harms \citep{barocas2017problem, crawford2017trouble, blodgett2020language}. Allocational harms concern the unfair distribution of resources, e.g.\ when a group is denied bank loans disproportionally by an automated system. Allocational harms tend to be more easily measured through standard fairness metrics like demographic parity \cite{dwork2012fairness} and equality of opportunity \cite{hardt2016equality}. Those do not directly apply to open-ended generation tasks, where we instead study representational harms, which arise when a system ``represents some social groups in a less favourable light than others, demeans them, or fails to recognise their existence altogether'' \citep{blodgett2020language}; the last case being the focus of our work. 

% comparative measures
\textit{Fairness measures for language generation}
usually define bias as differences between demographic groups \citep{sheng2021societal}. For example, \citet{dhamala2021bold} find that female pronouns are more likely to elicit positive text from an LLM than male pronouns. Similarly, \citet{huang2019reducing} compare different occupations, names and countries on produced sentiments. \citet{nangia2020crows} and \citet{nadeem2020stereoset} compare the probability of stereotypical and non-stereotypical sentences under a model in order to measure whether it encodes stereotypes against different demographic groups. Along the same lines, the WinoGender test \cite{rudinger2018gender} measures gender biases in co-reference resolution tasks. 
Taking a distributional view similar to our work, \citet{rae2021scaling} investigate biases in the context of occupation, however, they again compare predictions for different genders with each other. In general, such comparative bias tests are well-adapted by authors proposing new models \cite{touvron2023llama, rae2021scaling, hoffmann2022training, scao2022bloom}. Instead of comparing model predictions against each other, we compare model predictions to \textit{real world ground truth distributions} in order to quantify bias. 

% ground truth-based measures
\textit{Ground truth-based measures} are not commonly used as a metric for fairness but important when evaluating a model's truthfulness. \citet{petroni2019language} and \citet{lin2021truthfulqa} provide datasets of real world facts against which to benchmark LLMs' knowledge. Similar to our work, \citet{zhou2022richer} measure the frequency of country predictions, and how underprediction correlates with a country's GDP. Contrary to their count-based approach we propose a more fine-grained metric for erasure and extend the analysis to auto-regressive models. Unlike theirs, our erasure metric can be employed as a loss function for finetuning, to specifically mitigates erasure. \citet{liang2022holistic} propose a similar metric for erasure in the domains of gender and race. Like us, they compare model to ground truth distributions, though they measure a total variation distance where we use a KL-divergence based approach (see \S\ref{sec:method_erasure}). The authors assume uniform ground truth whereas we construct a domain specific distribution (see \S\ref{sec:method_ground_truth}). Lastly, unlike ours, their analysis does not cover any mitigation efforts.
Similar in spirit, geographical representativeness has been studied 
for text-to-image generation models \citep{rojas2022dollar, basu2023inspecting}.

\section{Method}\label{sec:method}

Our goal is to measure, and later mitigate, the extent 
to which large pretrained models underpredict some countries when generating language. We formalise this notion here. Note that while we are studying autoregressive models in this work, the methodology extends straightforwardly to masked models. Similarly, we are interested in measuring and reducing geographical erasure, 
but the analysis can be applied to other attributes where ground truth is available. For example, one could measure erasure with respect to age, ethnicity, religion or gender using the same formalism. 

\subsection{Obtaining Model Predictions}\label{sec:method_model_predictions}
Let $p$ be our language model over vocabulary $\Omega$. We consider open-ended generation tasks for autoregressive models. Such models predict the next token given previous ones, i.e.\ for a sequence of $L$ tokens $x^{1:L} \subset \Omega$ the probabilities factorise as 
\begin{align} \label{eq:autoregressive_model}
p(x^{1:L}) = \prod_{k=1}^{L-1} p( x^{k+1} | x^{1:k}).
\end{align}

We use pretrained models and condition on a short prompt, or context, of variable length $L$: $c=x^{1:L}$. Given this prompt, we compute the predictive distribution over a set of $M$ candidates $\{x_i\}_{i=1}^M = \mathcal{X} \subset \Omega$; see \S\ref{sec:method_ground_truth} for how these $M$ countries are chosen. 
For a candidate country $x_i \in \mathcal{X}$ we compute $p(x_i |c)$ as 
\begin{align} \label{eq:definition_p}
    p(x_i | c) = \frac{p(x_i, c)}{ p(c)} = \frac{p(x_i, c)}{\sum_{x \in \mathcal{X}} p(x, c) },
\end{align}
i.e., we compute $p($``I live in $x_i$''$)$ for all candidate countries $x_i$ and normalise. If a country is tokenised into multiple tokens, $x_i=x_i^{1:J}$, we multiply the probability of the $J$ subtokens according to \eqref{eq:autoregressive_model}. As before, superscript indicates position and subscript indicates the country name, e.g., $x_7 = \text{``Uganda''}$ is tokenised into $x_7^0 =$``U'', $x_7^1 =$``g'', $x_7^2 =$``anda''. As a consequence, $p(x_i | c)$ tends to be smaller for multi-token country names. Concerningly, \citet{zhou2022richer} show that this issue predominately impacts low GDP-countries.

Some countries in $\mathcal{X}$ are referred to by more than one name, e.g., ``UK'' and ``United Kingdom''. We disambiguate the countries using a list of alternative names\footnote{List of alternative country names from \url{https://en.wikipedia.org/wiki/List_of_alternative_country_names}, retrieved on Sept.~26th, 2023.}  to obtain the final $p(x_i | c) = \sum_{a \in \mathcal{A}} p(x_i^a|c)$ for all alternative names $x_i^a$.

In the following sections, we sometimes write $p(x_i|c) = p_i$, omitting the dependency on the prompt unless ambiguous. Note that we work directly on the model probabilities and discuss the impact on generated language in \S\ref{sec:limitations}.

\begin{figure*}
\includegraphics[width=\linewidth]{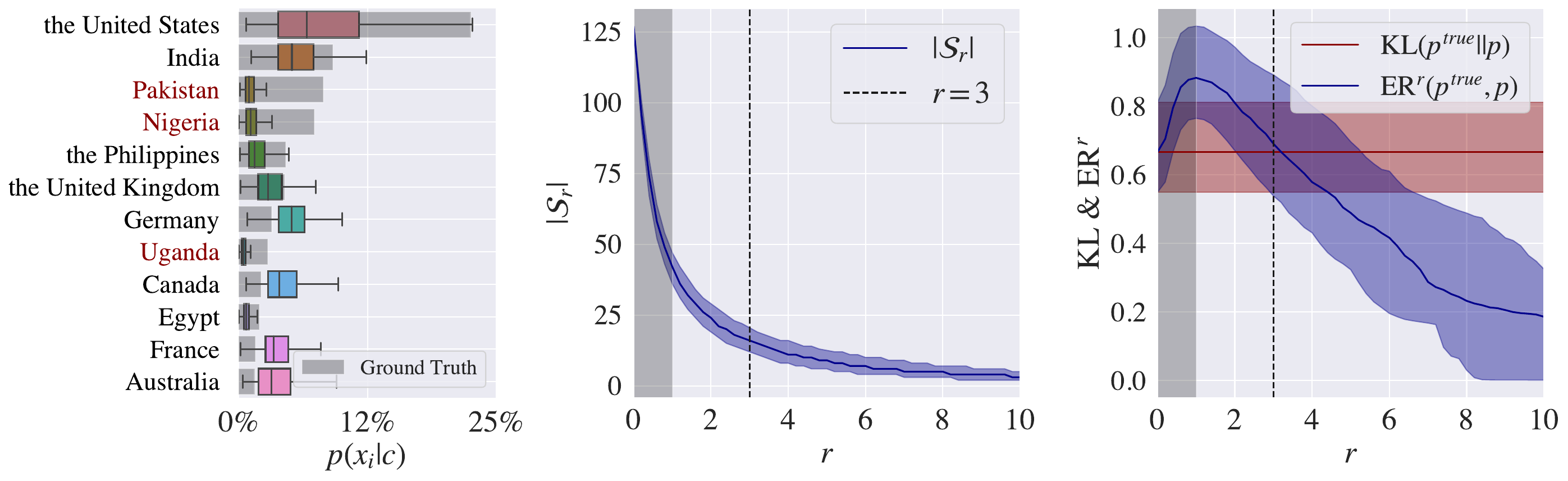}
\caption{\textbf{Understanding erasure.} \textit{Left:} OpenLLaMA, 7B vastly underpredicts the occurrence of Pakistan, Nigeria and Uganda. We plot country predictions given prompts $p(x_i|c)$ for different re-phrasings of the prompt ``I live in'' (boxplots) and ground truth (barplot, grey). Country names experiencing erasure ($x_i \in \mathcal{S}_3$, see \S\ref{sec:method_erasure}) are in red. We show the $12$ countries with the largest English speaking populations (in decrasing order). \textit{Middle: }Erasure set size $|\Sr|$ as a function of $r$ for OpenLLaMA, 7B. We plot the median (solid line) and $25^{th} - 75^{th}$ percentiles (blue shaded area) over different rephrasings (see \S\ref{sec:rephrasing}) of the same prompt. The dashed line marks $r=3$, the threshold we use in the experiments. This choice is further motivated in \S\ref{sec:properties}. \textit{Right:} Comparing ER$^r$ (blue) for different $r$ to the KL-divergence (red). We pick $r=3$, the integer value for which KL and ER$^r$ are the most similar.} \label{fig:example_predictions_and_r_param}
\end{figure*}

\subsection{Obtaining Ground Truth}\label{sec:method_ground_truth}
To measure erasure, we compare the generation distribution (given by equation~\ref{eq:definition_p}) to a ground truth distribution $\ptrue$ over the candidate countries, writing $\ptrue(x_i) = \ptrue_i$ as before. The ground truth is given by real world data, i.e., we compare our predictions to the actual population of country $x_i$. We adjust for the fact that our models are trained on English texts only by considering \textit{English speaking} populations as ground truth (see \S\ref{sec:limitations} for limitations of this approach). The number of English speakers per country is obtained from a Wikipedia list containing data for $M=127$ countries at the time of writing---we use these $127$ countries for our analysis.\footnote{From \url{https://en.wikipedia.org/wiki/List_of_countries_by_English-speaking_population}, retrieved on Sept.~26th, 2023.} Unlike the model predictions $p(x_i|c)$, the ground truth $\ptrue(x_i)$ is prompt-independent. We will generalise model predictions to be prompt-independent as well by marginalising prompts in \S\ref{sec:rephrasing}. See Figure \ref{fig:example_predictions_and_r_param} (left) for an example of model predictions and ground truth.

\subsection{Measuring Erasure}\label{sec:method_erasure}
With these prerequisites in place, we can now formalise erasure using the relationship of predictive distribution and ground truth. 

\begin{definition}[Erasure Set]
For a ratio threshold $r > 1$ we define the erasure set under model $p$, ground truth $\ptrue$ and prompt $c$ as 
\begin{equation} \label{eq:S_r}
    \Sr^{c} = \left\{ x_k: \frac{\ptruedef(x_k)}{p(x_k| c) } > r \right\}.
\end{equation}
\end{definition}

In the example in Figure \ref{fig:example_predictions_and_r_param} (left), we prompt the OpenLLaMA model with different versions of the prompt ``I live in'' and aggregate the predictions (see \S\ref{sec:rephrasing} for rephrasing and aggregation). We then compute the erasure set for $r=3$, i.e.\ countries that are three times more prevalent in the ground truth than in our predictions. We obtain $\mathcal{S}_{3}=$\{Pakistan,
Nigeria,
Uganda,
Bangladesh,
Iraq,
Madagascar and
Eswatini\}. A simple metric of erasure is the size of the erasure set $|\Sr|$ for a user-specified $r$, here, $|\mathcal{S}_{3}|=7$.

$|\Sr|$ measures \textit{how many} countries are ``erased'' (underrepresented by at least factor $r$). To obtain a more fine grained numerical evaluation we measure \textit{by how much} they are underrepresented compared to ground truth by reporting the following metric.

\begin{definition}[Erasure]
Erasure w.r.t.\ ground truth $\ptrue$ at threshold $r$ is defined as 
\begin{align} \label{eq:ER^r}
\emph{ER}^r(p^\emph{true}, p) &= \sum_{i \in \Sr} \ptruedef_i \log \left( \frac{\ptruedef_i}{ p_i} \right).
\end{align} 
\end{definition}

\subsection{Properties of ER$^r$}\label{sec:properties}
A careful conceptualisation of any proposed fairness metric is crucial \cite{schwobel2022long, blodgett2020language}. We motivate our definition of ER$^r$ here. 
Firstly, if $p = \ptrue$ then EB$^r(\ptrue, p) =0$ for all $r$; i.e., no erasure occurs when the distributions match. 
Secondly, unlike the total variation distance suggested in \citet{liang2022holistic}, we want our metric to be sensitive to \textit{relative} rather than absolute errors, so that countries with small populations are also taken into account. Hence we report (log-)ratios in the definition of ER$^r$ \eqref{eq:ER^r}.  On the other hand, while we believe this sensitivity to less-populated countries is important, we do acknowledge that underpredicting big ground truth populations is particularly harmful as it impacts a lot of users. Thus, we weight the log-ratios by the ground truth probabilities $\ptrue_i$. 

A third factor in our choice of metric is the close relation of \eqref{eq:ER^r} and the KL-divergence KL$(\ptrue || p)$. ER$^r$ is an additive component of the KL-divergence: 
 \begin{align} \label{eq:KL_terms}
 \text{KL}(\ptrue || p) = \sum_{i \in \Sr} & \ptrue_i \log \left( \frac{\ptrue_i}{ p_i} \right) +   \\
 & \sum_{i \in \mathcal{X} \setminus \Sr}  \ptrue_i \log \left( \frac{\ptrue_i}{ p_i} \right). \nonumber
 \end{align}

This close relation to a well-defined divergence measure 
allows for theoretical analysis and helps practitioners 
build on existing intuitions. 

\textbf{The choice of $\mathbf{r}$} is a crucial hyperparamter, as $|\Sr|$ and ER$^r$ are defined in terms of $r$. We discuss the impact here and visualise it in Figure \ref{fig:example_predictions_and_r_param} (middle and right).
For small values of $r$, we include all terms in \eqref{eq:KL_terms}, i.e., $$\lim_{r\to0} \Sr = \mathcal{X} \text{ and } \lim_{r\to0} \text{ER}^r(\ptrue, p) = \text{KL}(\ptrue || p).$$
For larger values of $r$, we instead have 
$$\lim_{r\to\infty} \Sr = \emptyset \text{ and } \lim_{r\to \infty} \text{ER}^r(\ptrue, p) = 0.$$
See Figure \ref{fig:example_predictions_and_r_param} (right) for this relationship. Since we want to measure erasure or underprediction, we study cases where $\ptrue > p$, i.e., for values $r  > 1$.\footnote{There is a degree of symmetry in our measurement: being probability distributions, $\ptrue$ and $p$ sum to one. Thus, when $\ptrue > p$ for $\Sr$, there are other countries for which the opposite is true, i.e.\ that are overpredicted. In general, we believe underprediction to be more likely to cause harms than overprediction (see \S\ref{sec:intro}), hence we focus on measuring erasure.}
We pick $r$ to be an integer such that $\text{ER}^r (\ptrue, p) \approx \text{KL}(\ptrue || p)$, that is $r=3$ in the experiment in Fig.~\ref{fig:example_predictions_and_r_param} (right). We find that this value is the same across all our models (see Appendix \ref{sec:appendixA}), so we choose $r=3$ globally. This choice of $r$ is based on a mathematical heuristic. An alternative way of choosing this parameter might be implied by legal or ethical constraints. For example, a guideline on adverse impact by the \citet{us1979questions} defines ``a substantially different rate of selection'' at $80\%$. In this labour market use case, $r = 1/ 0.8 = 1.25$ would be the corresponding hyperparameter.

\textbf{Differentiability} is an important property of our metric since we want to use it for finetuning LLMs in \S\ref{sec:exp4}. For fixed $r$, ER$^r$ is differentiable almost everywhere (with respect to the network weights). Singularities occur at those points that add new countries to the erasure set $\mathcal{S}_r$ in \eqref{eq:S_r}, i.e., weights such that $\ptrue_k = p_k$ for any country $k$.

\subsection{Prompt Rephrasing}\label{sec:rephrasing}
The erasure set definition in \eqref{eq:S_r}, and consequently the notion of erasure in \eqref{eq:ER^r} are prompt-dependent. However, we are interested more generally in the model's world knowledge rather than its completion of a specific prompt. Hence, we would like to aggregate the effect over all prompts encoding the meaning $\mathcal{M}=$``home country'', by using the following marginal distribution:
\begin{align}  \label{eq:marginal_pred}
    p(x| \mathcal{M}) = \int p(x| c) p(c | \mathcal{M})\text{d}c.
\end{align}

The relationship between a prompt $c$ and its meaning $\mathcal{M}$ is complex, hence computing \eqref{eq:marginal_pred} is intractable. Here, we will rely on simple, pragmatic techniques to semi-automatically construct a set of sample prompts $\mathcal{D} \sim P(c | \mathcal{M})$ from a seed prompt $\tilde{c}$. We rephrase $\tilde{c}$ while preserving its meaning to generate additional prompts. This is common practice: \citet{jiang2020can} use mining- and translation-based paraphrasing methods while \citet{romano2006investigating} rely on templates for paraphrasing. In light of recent advances in LLMs, another way to automatically rephrase prompts is by using a model that has been finetuned for paraphrasing \cite{niu2020unsupervised}. Even simpler, we use an off-the-shelf model by prompting ChatGPT to rephrase the $\tilde{c}=$``I am from'' seed prompt.\footnote{Accessed via \url{https://chat.openai.com/}.} After manually removing irrelevant prompts we obtain $16$ base formulations. We further expand the set of prompts by replacing sentence subjects. For example, we expand ``I live in'' into \{``You live in'', ``He lives in'',  ``She lives in'', ...,\}, producing a total of $|\mathcal{D}| = 955$ prompts. Details and a list of all prompts can be found in Appendix \ref{sec:appendixB}. We use the dataset of $955$ prompts to approximate the marginal in \eqref{eq:marginal_pred} assuming different prior probabilities $p(c | \mathcal{M})$ as follows: 

\paragraph{(1) Uniform prompt distribution:}
\begin{align}
p( c | \mathcal{M}) = \frac{1}{|\mathcal{D}|}, \text{ then}
\end{align}
\begin{align} p(x|\mathcal{M})\!\approx\!\frac{1}{|\mathcal{D}|} \sum_{c \in \mathcal{D}} p(x| c)\!=\!p^{\text{agg\_uni}}(x | \mathcal{M})
 \label{eq:uniform_agg}. \end{align}

\paragraph{(2) Model-induced prompt distribution:}
\begin{align}
p( c | \mathcal{M}) =
\frac{p(c)}{\sum_{c \in \mathcal{D}} p(c)}
\end{align} 
where $p(c)$ is the probability given by the autoregressive language model \eqref{eq:autoregressive_model}. In this case, 
\begin{align}  \label{eq:model_based_agg} 
p(x|\mathcal{M}) &\approx\sum_{c \in \mathcal{M}} p(x| c)  p(c |\mathcal{M}) \nonumber \\
& =p^{\text{agg\_model}}(x | \mathcal{M}).
\end{align}

\section{Experiments}\label{sec:exp}
In this section, we show the existence of geographical erasure across different LLMs and different prompt wordings (\S\ref{sec:exp1}). We highlight the consistency of erased countries across models (\S\ref{sec:erasure_who}) and investigate the impact of model size on erasure (\S\ref{sec:exp2}). We identify possible causes of erasure (\S\ref{sec:exp3}) and explore a mitigation strategy (\S\ref{sec:exp4}).

\textbf{The models under consideration} are GPT2 \cite{radford2019language}, $117$M, $345$M, $774$M and $1.6$B weight versions, GPT-Neo \cite{gpt-neo}, $125$M, $1.3$B and $2.7$B weight versions, GPT-NeoX, $20$B weights \cite{black2022gpt} and open source reproductions of the LLaMA model \cite{touvron2023llama, openlm2023openllama, together2023redpajama}, $3$B and $7$B weights. We obtain all implementations from HuggingFace.\footnote{Via \url{https://huggingface.co/docs/transformers}.}

\begin{figure*}[t]
\includegraphics[width=\linewidth]{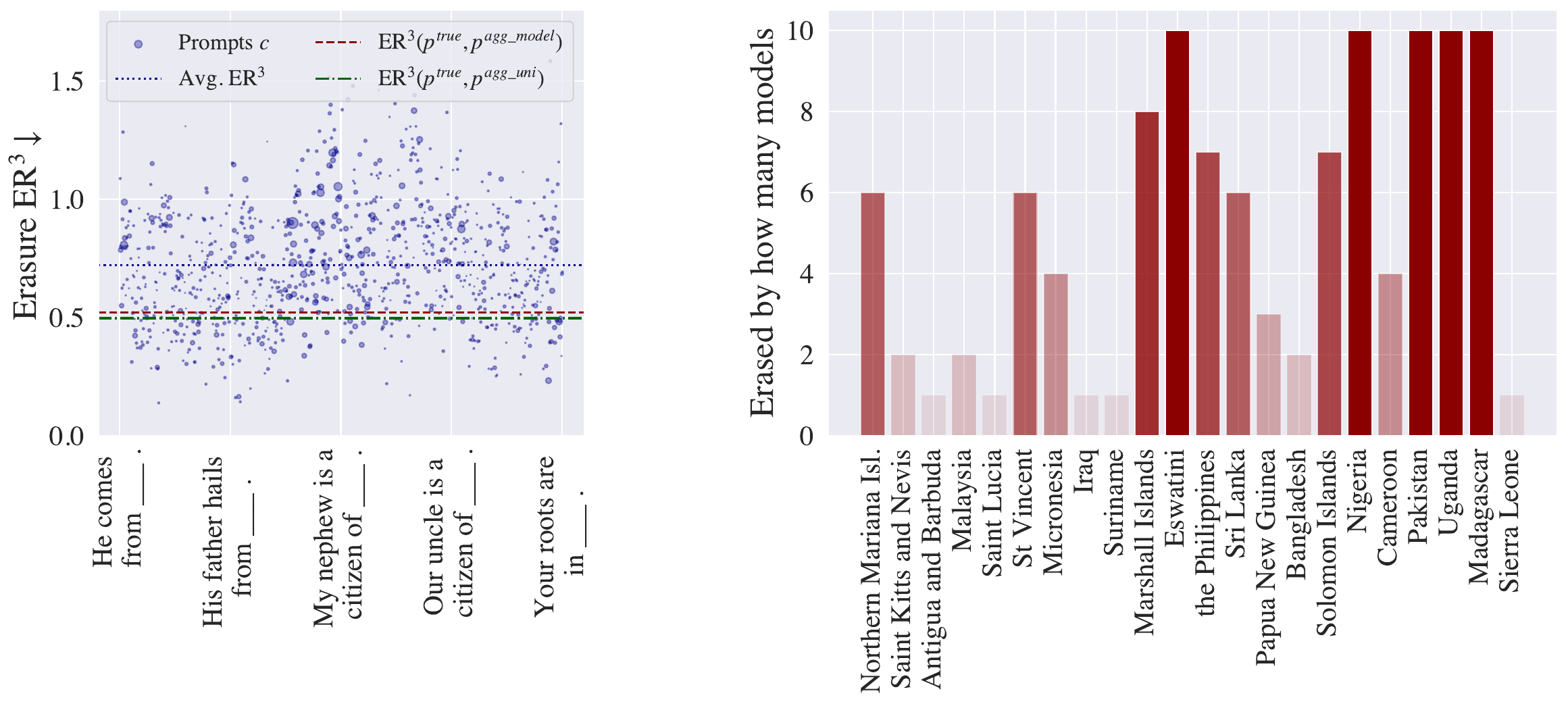}
\caption{\textbf{Geographical erasure occurs for all prompt rephrasings, and many countries experience erasure consistently under all models.} \textit{Left:} OpenLLaMA, 7B results for $955$ individual prompts (blue dots) along the x-axis, with some example prompts as axis labels. We also plot ER$^r$ in aggregate: The blue line is the average over individual prompts $\frac{1}{C}\sum_c\text{ER}^3(\ptrue, p(\cdot | c)$),  green is the uniform aggregate ER$^3(\ptrue, p^{\text{uni\_agg}})$ \eqref{eq:uniform_agg} and red is the model-induced aggregate ER$^3(\ptrue, p^{\text{model\_agg}})$ \eqref{eq:model_based_agg}. Size of dots corresponds to the probability assigned to the respective prompt under the model. The gap between blue and red/green aggregations is explained in \S\ref{sec:rephrasing}. \textit{Right:} Of the $M=127$ countries, $105$ do not experience erasure at $r=3$ for any of the models. For the remaining $22$, we plot model counts here. Bars are coloured according to counts and sorted by GDP per capita (decreasing from left to right). We use aggregated predictions according to Equation \ref{eq:model_based_agg}.} \label{fig:exp1_exp2}
\end{figure*}

\subsection{Impact of Prompt Wording}\label{sec:exp1}
We start by investigating how dependent erasure is on the exact phrasing of the prompt. We prompt the models with rephrased versions of ``I live in'' (see \S\ref{sec:rephrasing}) and compute erasure $\text{ER}^3(\ptrue, p( \cdot | c))$ for each prompt $c$. In Figure~\ref{fig:exp1_exp2} (left), we plot the (i) erasure for individual prompts (dots); (ii) the average erasure $\frac{1}{C}\sum_c\text{ER}^3(\ptrue, p(\cdot | c)$) denoted by the blue dotted line; (iii) erasure for the uniform marginal distribution from \eqref{eq:uniform_agg} using a green dash-dotted line; and (iv) erasure for the model-induced marginal distribution from \eqref{eq:model_based_agg} as a red dashed line. The size of the blue dots indicates $p(c | \mathcal{M})$.% under the model. 

The magnitude of erasure  $\text{ER}^3(\ptrue, p( \cdot | c))$ differs across the phrasings $c$, however, \emph{erasure exists in all versions} (that is, $\text{ER}^3>0$ with p-value $\ll0.01$). 
We note that erasure under the aggregate distribution is smaller than the average erasure ($\text{ER}^3(\ptrue, \pagguni) < \frac{1}{C}\sum_c\text{ER}^3(\ptrue, p(\cdot | c)$ in  Figure \ref{fig:exp1_exp2} (left)). This follows from Jensen's inequality (see Appendix \ref{sec:appendixC} for details).
Throughout the remainder of the paper, we will report the aggregates from \eqref{eq:uniform_agg} and \eqref{eq:model_based_agg} along with boxplots of ER$^3$ to account for the variance due to rephrasings.

\subsection{Who is Experiencing Erasure?}\label{sec:erasure_who}
We evaluate whether the same countries experience erasure under all the examined $10$ models, and what characterises these countries. Out of the $M=127$ countries under analysis, $105$ do not experience erasure at $r=3$ for any of the models. For the remaining $22$ nations, Figure \ref{fig:exp1_exp2}~(right) shows the number of models by which they are erased. Worryingly, Eswatini, Nigeria, Pakistan, Uganda and Madagascar experience erasure under \textit{all} $10$ analysed models. The x-axis in Figure~\ref{fig:exp1_exp2}~(right) is ordered by GDP per capita, in decreasing order from left to right.\footnote{Data from \url{https://en.wikipedia.org/wiki/List_of_countries_by_GDP_(nominal)} on Sept.~26th, 2023.}

\begin{figure*}[t]
\includegraphics[width=\textwidth]{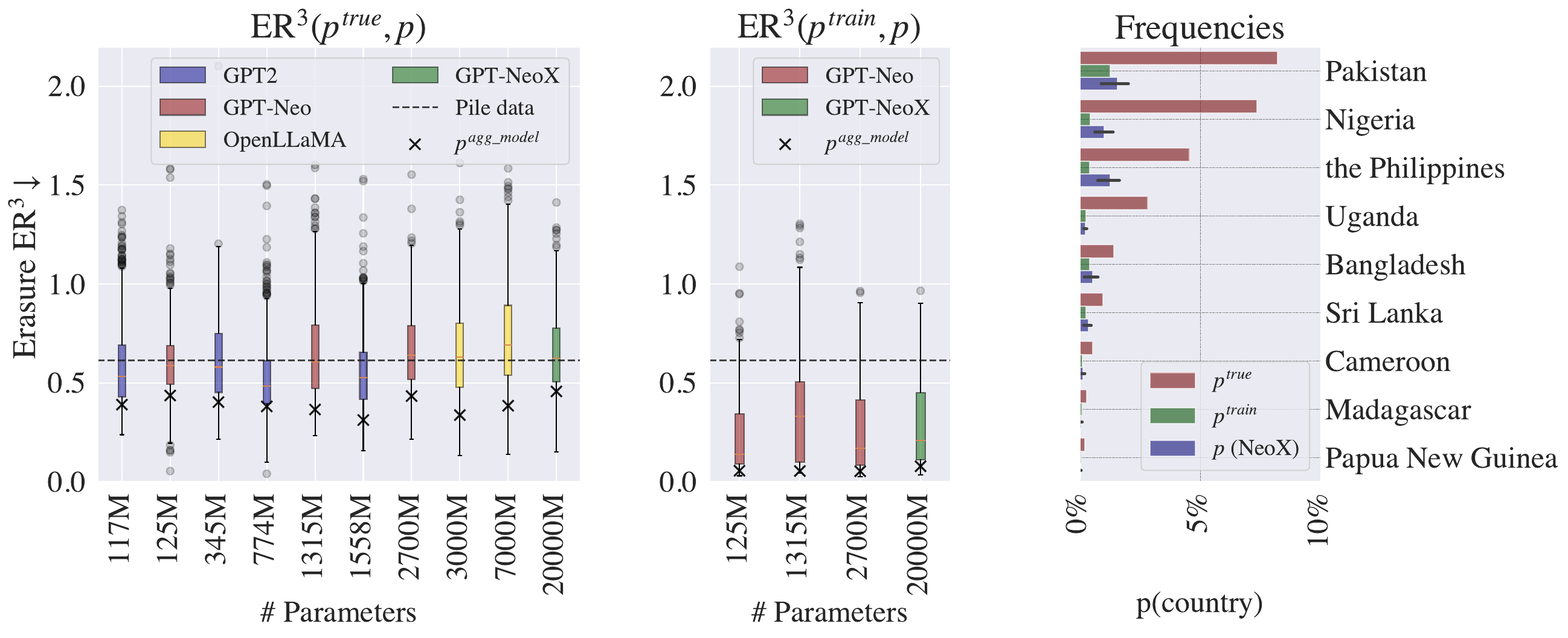}
\caption{\textbf{Erasure in models closely matches distribution of country mentions in training data.} \textit{Left:}  Geographical erasure or GPT-type models of different sizes. Size is on the x-axis (axis not to scale). Blue models are GPT2, red models are GPT-Neo, yellow are OpenLLaMA and the green model is GPT-NeoX, crosses below each box plot are aggregated results over different prompts (see Eq.~\ref{eq:model_based_agg}). The dashed line indicates ER$^r(\ptrue, \ptrain)$. \textit{Middle:} Geographical erasure for GPT-type models of different sizes, assuming training frequency as the ground truth (instead of world population). \textit{Right:} Ground truth (red) and Pile training data (green) distributions compared to GPT-NeoX predictions (blue) on countries in the erasure set $\mathcal{S}_3$ (of GPT-NeoX w.r.t.\ ground truth).} \label{fig:model_size_pile}
\end{figure*}

\subsection{Impact of Model Size}\label{sec:exp2}

Related work \citep{lin2021truthfulqa, rae2021scaling, nadeem2020stereoset} reports mixed results on the relationship between model size and bias. On the one hand, \citet{lin2021truthfulqa} report that on the TruthfulQA benchmark, ``[l]arger models are less truthful''. This is because large models surface the common human misconceptions that the questions are designed to elicit. Such misconceptions are likely present in the training data which the larger models match more faithfully. Similarly, \citet{nadeem2020stereoset} find that the larger models exhibit more stereotyping, again this is probably because they match stereotypes in the training data more closely. On the other hand, \citet{rae2021scaling} ``do not see a consistent correlation between model size and bias'' in their tests for gender-occupation bias. 

We visualise the extent of geographical erasure with varying model sizes in Figure \ref{fig:model_size_pile} (left). 
Like \citet{rae2021scaling}, we do not find model size to have a big impact. We hypothesise that even the smaller models closely mimic the frequency distribution (of country mentions) in the training corpus, similar to \citet{rae2021scaling}'s experiment. We believe that this is not the case in the test by \citet{lin2021truthfulqa} and \citet{nadeem2020stereoset}, because their tests go much beyond unigram frequencies, and smaller models do not exhibit such subtle biases. We explore the relationship of data bias and model bias below.

\begin{figure*}
\centering
\includegraphics[width=\textwidth]{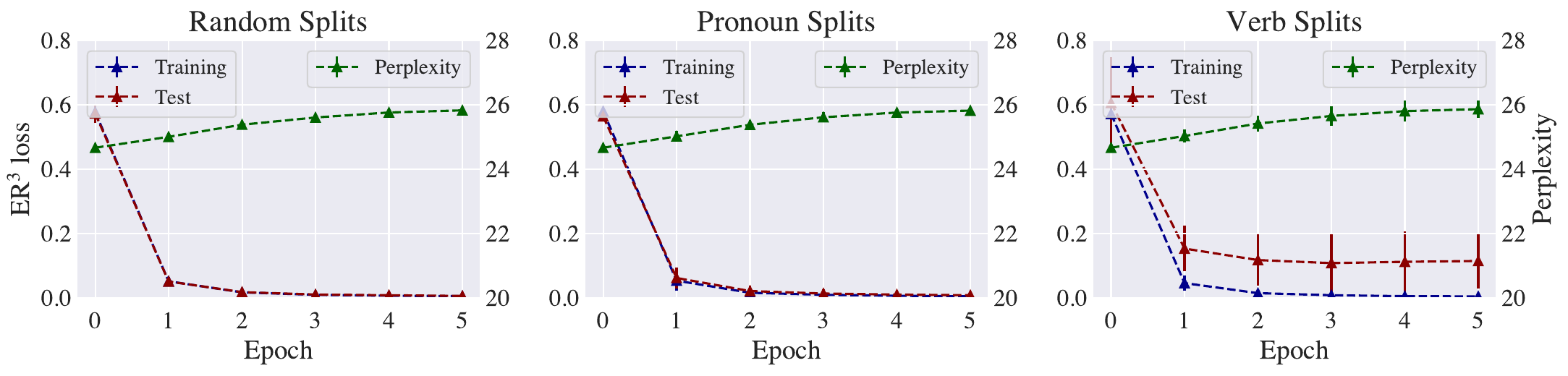} 
\caption{\textbf{Finetuning effectively alleviates erasure.} We plot average ER$^r$ on training (blue) and test (red) set prompts during $5$ epochs of finetuning of the GPT2-small model. Error bars indicate minima/maxima over $5$ folds.
} \label{fig:mitigation_exp}
\end{figure*}

\subsection{Impact of Training Data}\label{sec:exp3}

We hypothesise that training data is an important factor for erasure: models underpredict countries which appear in the data infrequently compared to their population. To study the relationship between training data bias and model bias, we extract the distribution of country mentions in the training data. 
We consider the Pile dataset~\cite{DBLP:Pile} used to pre-train the GPT-Neo LLMs analysed in this study.
To determine the probability of occurrence in the training data $\ptrain(x)$, we compute the number of times each country $x$ is mentioned in the dataset, i.e., $\ptrain(x) \propto \textit{\# mentions of x}$. These mention counts are weighted by the number of training epochs this document was included while training (dataset weights $w_d$ from \citet{DBLP:Pile}). We account for alternative country names as described in \S\ref{sec:method_model_predictions}. Thus, the final formula becomes 
\begin{align}
\ptrain(x) \propto \sum_{d \in \text{datasets}} w_d\sum_{a \in \mathcal{A}} \textit{\#} x^a \in d,
\end{align}

where $\mathcal{A}$ represents a set of alternative country names of a given country and \# represents counts. Once all the counts are gathered, the results are normalised to determine the final values of $\ptrain(x)$, which we compare to the outputs of LLMs.

Specifically, we compare ground truth $\ptrue(x)$, training data $\ptrain(x)$ and GPT-NeoX predictions $p(x)$ for countries $x \in \mathcal{S}_3$ (Figure~\ref{fig:model_size_pile}, right). We see that countries experiencing erasure are indeed underrepresented in the training data, and the prediction probabilities 
of these countries 
are similar to their frequency distribution in the training corpus ($\ptrue(x) \gg p(x) \approx \ptrain(x))$.

We then compute erasure against the training data, ER$^r(\ptrain, p)$, i.e., considering the ground truth to be $\ptrain$ (Figure~\ref{fig:model_size_pile}, middle). 
We find that erasure values in this case 
are considerably lower. For instance
ER$^r(\ptrain, \paggmodel)=0.08$ for GPTNeoX
compared to ER$^r(\ptrue, \paggmodel) = 0.46$, i.e., erasure using the world population (Figure~\ref{fig:model_size_pile}, left).
This indicates that the GPTNeoX family of LLMs mimic the training distribution (of country mentions). Furthermore, we find that the erasure score of the training data compared to ground truth,  ER$^r(\ptrue, \ptrain)$, is itself $0.46$, which closely matches the erasure for models trained on this data. 
The high correlation between data bias and model bias suggests the composition of training data is a key source of erasure in the investigated LLMs.

\subsection{Mitigation}\label{sec:exp4}

In this section, we explore finetuning as a strategy to mitigate erasure. We perform gradient updates on a pretrained GPT2 model to minimise the erasure loss ER$^3(\ptrue, p)$ on the training data given by prompt data set $\mathcal{D}$. We note that our finetuning strategy differs from the related approach from \S6.2 of \citet{zhou2022richer} in that we have formulated a loss function which allows us to perform supervised finetuning. \citet{zhou2022richer} instead continue training the model using the standard masked language modelling loss with augmented data related to underpredicted countries. 
% TODO: say why supervised is expected to work better, and ideally cite a reference

We use the AdamW \cite{loshchilov2017decoupled} optimiser with learning rate $3e-5$ and train for an additional $5$ epochs (including one epoch of warmup under a linear schedule). We find that due to our loss function's direct dependency on the logits and the re-normalisation of probabilities over $\mathcal{X}$ \eqref{eq:definition_p} our finetuning strategy works best for deterministic models, hence we set dropout rates to $0$ for embeddings, encoder, pooling and attention layers. For finetuning, we update the bias terms only, following \citet{zaken2021bitfit}. This is a memory efficient strategy that is expected to work particularly well in settings with constant outputs (we want the generated distribution for \textit{all} our prompts to match the ground truth), while not impacting the general language modelling abilities. We evaluate whether the language modelling abilities deteriorate by measuring perplexity on Wikitext-2-v1 \cite{merity2016pointer} before and after every epoch of finetuning.

To measure how well our finetuning strategy generalises, we compare three different ways of performing train-test splits of our $955$ prompts. These include \textit{Random} partitioning: we randomly split the prompts into $75\%$ training and $25\%$ test data; \textit{Pronouns}: we split the prompts based on the pronouns they contain, e.g.\ all prompts containing ``she'', ``you'', ``we'' and ``they'' are in the training set, ``I'' and ``'he' in the test set; \textit{Verbs}: we divide along verb groups, e.g.\ prompts containing ``to live in'' and `to be a citizen'' of are in the training set, ``to reside in'' is in the test set. These three setups 
require increasing 
levels of generalisation.

For all three setups, we repeat the experiment on $5$ different folds and plot the results in Figure \ref{fig:mitigation_exp}. We find that our finetuning strategy is  effective: the average erasure $\frac{1}{|\mathcal{D}|} \sum_{c \in \mathcal{D}} \text{ER}^3 (\ptrue, p(\cdot | c))$ is small after $5$ epochs of finetuning, both on training (blue) and test data (red). 
The model generalises well in the random case (Figure \ref{fig:mitigation_exp}, left) and to new pronouns (Figure \ref{fig:mitigation_exp}, middle).
As expected, \textit{verb splits} are the most challenging for our model, where we see that the erasure values decrease but not as much as we see in other splits (Figure \ref{fig:mitigation_exp}, right). In all cases, we see only a small deterioration in language modelling performance, as indicated by an approximate 5\% increase in the perplexity (The green lines in the plots of Figure~\ref{fig:mitigation_exp} correspond to perplexity). We compare this successful mitigation strategy to alternatives in Appendix \ref{sec:appendixD}.

\section{Conclusion}\label{sec:conclusion}

We motivated 
the need for large language models 
to be more geographically inclusive---which remains to be an overlooked aspect of inclusive model development.
Specifically, we studied and formalised a notion of geographical erasure, which captures the 
countries 
that are underpredicted
and the extent to which they are underpredicted. 
We discussed how our formulation captures many desirable 
properties.
In our experiments, we found clear 
instances of geographical erasure, which were 
consistently 
observed across $10$ different language models. 
Perhaps unsurprisingly,
the 
output probabilities 
of language models 
closely follow the
frequencies
of country mentions
in the training corpora, a likely cause of erasure. 
We examined a finetuning-based mitigation
strategy 
and found it to be effective in alleviating 
erasure.

% Note that this section does NOT count towards 8p page limit
\section{Limitations}\label{sec:limitations}
\paragraph{Languages considered.} We limit our analysis to models trained on English texts, and hence we prompt them in English only. Our methodology extends to other languages straightforwardly. For example, to replicate the geographical experiment with a Spanish language model, one would auto-generate Spanish prompts (or translate the English ones from Appendix \ref{sec:appendixB}).

The language (of prompts) used to 
analyse erasure should be accounted for 
while collecting ground truth data: 
for instance, English speaking countries  are expected to have higher probability conditioned on ``I live in'', and similarly Spanish speaking countries conditioned on ``Vivo en'' are likely to have higher probabilities. In our work, we factor this by considering English speaking populations as ground truth in \S\ref{sec:exp} (and one would proceed accordingly for a model in a different language). 

\paragraph{Difficulty in obtaining ground truth.} 
Language specific ground truth data is less reliable and harder to obtain than raw population counts. Such statistics are often self-reported and the level of proficiency differs dramatically across regions, especially since the numbers include second language speakers.\footnotemark[2] Since we only measure erasure for countries where $\ptrue_i$ is available, the availability of language specific ground truth data is itself a biasing factor. This is evident from Figure~\ref{fig:teaser}, which depicts how the lack of ground truth data predominantly affects
central African regions.

\paragraph{Knowledge encoding vs.\ language generation.} 
Our erasure metric is based on country probabilities given a prompt $p(x_i | c)$. These probabilities can be interpreted as knowledge encoded by the model. When generating text, the model probabilities are used to sample next tokens. Sampling (or decoding) can be performed using different strategies, e.g.\ greedy or beam search \cite{klein2017opennmt} to maximise probabilities, or top-k \cite{fan2018hierarchical} and top-p \cite{holtzman2019curious} sampling strategies to generate more diverse outputs. 

Our work does not analyse the effect of the decoding mechanism since we work directly on $p(x_i | c)$ instead of generated text. This is not uncommon in prior work, likelihood-based methods such as perplexity or cross-entropy are a customary way to evaluate language modelling abilities, also in modern LLMs \cite{radford2019language}. 

Compared to evaluating the full generation pipeline, erasure (as defined in Equation \eqref{eq:ER^r}) can be thought of as a lower bound to erasure under sampling: instead of considering the full predictive distribution, the above sampling mechanisms only consider high-probability candidates, erasing low-probability countries to even larger degrees. 

\paragraph{Causes of erasure.} 
Our analysis covers two potential sources for erasure: training data and model size. Model bias is commonly explained by data bias (e.g.~\citet{bender2021dangers,schwobel2022learned} and \citet{buolamwini2018gender}). In our work, we have not experimentally established the cause, our experiments instead indicate a high correlation of model biases and data biases in \S\ref{sec:exp3}, suggesting that data is a likely source of erasure. Data, however, is not the only biasing factor. Model architecture and training paradigm determine how the data is used by the model. Hence, they determine whether data bias is mitigated or exacerbated \citep{hooker2021moving}. We examine the impact of model size and find that it has little to no impact on geographical erasure ( \S\ref{sec:exp2})). Examining the impact of other factors on erasure is left to future work.

\section*{Acknowledgements}

This work greatly benefited from discussions with Mansi Gupta, Luca Franceschi, Bilal Zafar, Gianluca Detommaso, Martin Wistuba and Prabhu Teja Sivaprasad. We also appreciate the feedback on our work by Navreet Kaur and anonymous EMNLP reviewers. 
DP acknowledges Adobe Inc.~and Kotak IISc AI-ML Centre (KIAC) for their support. % towards the research of his group.

\bibliography{anthology, custom}

\begin{thebibliography}{52}
\expandafter\ifx\csname natexlab\endcsname\relax\def\natexlab#1{#1}\fi

\bibitem[{Barocas et~al.(2017)Barocas, Crawford, Shapiro, and
  Wallach}]{barocas2017problem}
Solon Barocas, Kate Crawford, Aaron Shapiro, and Hanna Wallach. 2017.
\newblock The problem with bias: Allocative versus representational harms in
  machine learning.
\newblock In \emph{9th Annual conference of the special interest group for
  computing, information and society}.

\bibitem[{Basu et~al.(2023)Basu, Babu, and Pruthi}]{basu2023inspecting}
Abhipsa Basu, R~Venkatesh Babu, and Danish Pruthi. 2023.
\newblock Inspecting the geographical representativeness of images from
  text-to-image models.
\newblock \emph{Conference on Computer Vision and Pattern Recognition (CVPR)}.

\bibitem[{Bender et~al.(2021)Bender, Gebru, McMillan-Major, and
  Shmitchell}]{bender2021dangers}
Emily~M Bender, Timnit Gebru, Angelina McMillan-Major, and Shmargaret
  Shmitchell. 2021.
\newblock On the dangers of stochastic parrots: Can language models be too big?
\newblock In \emph{Proceedings of the 2021 ACM conference on fairness,
  accountability, and transparency}, pages 610--623.

\bibitem[{Black et~al.(2022)Black, Biderman, Hallahan, Anthony, Gao, Golding,
  He, Leahy, McDonell, Phang et~al.}]{black2022gpt}
Sid Black, Stella Biderman, Eric Hallahan, Quentin Anthony, Leo Gao, Laurence
  Golding, Horace He, Connor Leahy, Kyle McDonell, Jason Phang, et~al. 2022.
\newblock Gpt-neox-20b: An open-source autoregressive language model.
\newblock \emph{arXiv preprint arXiv:2204.06745}.

\bibitem[{Black et~al.(2021)Black, Gao, Wang, Leahy, and Biderman}]{gpt-neo}
Sid Black, Leo Gao, Phil Wang, Connor Leahy, and Stella Biderman. 2021.
\newblock \href {https://doi.org/10.5281/zenodo.5297715} {{GPT-Neo: Large Scale
  Autoregressive Language Modeling with Mesh-Tensorflow}}.

\bibitem[{Blank(2017)}]{blank2017digital}
Grant Blank. 2017.
\newblock The digital divide among twitter users and its implications for
  social research.
\newblock \emph{Social Science Computer Review}, 35(6):679--697.

\bibitem[{Blodgett et~al.(2020)Blodgett, Barocas, Daum{\'e}~III, and
  Wallach}]{blodgett2020language}
Su~Lin Blodgett, Solon Barocas, Hal Daum{\'e}~III, and Hanna Wallach. 2020.
\newblock Language (technology) is power: A critical survey of" bias" in nlp.
\newblock \emph{arXiv preprint arXiv:2005.14050}.

\bibitem[{Bolukbasi et~al.(2016)Bolukbasi, Chang, Zou, Saligrama, and
  Kalai}]{bolukbasi2016man}
Tolga Bolukbasi, Kai-Wei Chang, James~Y Zou, Venkatesh Saligrama, and Adam~T
  Kalai. 2016.
\newblock Man is to computer programmer as woman is to homemaker? debiasing
  word embeddings.
\newblock \emph{Advances in neural information processing systems}, 29.

\bibitem[{Brown et~al.(2020)Brown, Mann, Ryder, Subbiah, Kaplan, Dhariwal,
  Neelakantan, Shyam, Sastry, Askell et~al.}]{brown2020language}
Tom Brown, Benjamin Mann, Nick Ryder, Melanie Subbiah, Jared~D Kaplan, Prafulla
  Dhariwal, Arvind Neelakantan, Pranav Shyam, Girish Sastry, Amanda Askell,
  et~al. 2020.
\newblock Language models are few-shot learners.
\newblock \emph{Advances in neural information processing systems},
  33:1877--1901.

\bibitem[{Buolamwini and Gebru(2018)}]{buolamwini2018gender}
Joy Buolamwini and Timnit Gebru. 2018.
\newblock Gender shades: Intersectional accuracy disparities in commercial
  gender classification.
\newblock In \emph{Conference on fairness, accountability and transparency},
  pages 77--91. PMLR.

\bibitem[{Center(2021)}]{pew_research_internet_access}
Pew~Research Center. 2021.
\newblock \href
  {https://www.pewresearch.org/internet/fact-sheet/internet-broadband/}
  {Internet/broadband fact sheet}.

\bibitem[{Chowdhery et~al.(2022)Chowdhery, Narang, Devlin, Bosma, Mishra,
  Roberts, Barham, Chung, Sutton, Gehrmann et~al.}]{chowdhery2022palm}
Aakanksha Chowdhery, Sharan Narang, Jacob Devlin, Maarten Bosma, Gaurav Mishra,
  Adam Roberts, Paul Barham, Hyung~Won Chung, Charles Sutton, Sebastian
  Gehrmann, et~al. 2022.
\newblock Palm: Scaling language modeling with pathways.
\newblock \emph{arXiv preprint arXiv:2204.02311}.

\bibitem[{Computer(2023)}]{together2023redpajama}
Together Computer. 2023.
\newblock \href {https://github.com/togethercomputer/RedPajama-Data}
  {Redpajama-data: An open source recipe to reproduce llama training dataset}.

\bibitem[{Crawford(2017)}]{crawford2017trouble}
K~Crawford. 2017.
\newblock The trouble with bias—nips 2017 keynote—kate crawford\# nips2017.
\newblock \emph{The Artificial Intelligence Channel. https://www. youtube.
  com/watch}.

\bibitem[{Devlin et~al.(2018)Devlin, Chang, Lee, and
  Toutanova}]{devlin2018bert}
Jacob Devlin, Ming-Wei Chang, Kenton Lee, and Kristina Toutanova. 2018.
\newblock Bert: Pre-training of deep bidirectional transformers for language
  understanding.
\newblock \emph{arXiv preprint arXiv:1810.04805}.

\bibitem[{Dhamala et~al.(2021)Dhamala, Sun, Kumar, Krishna, Pruksachatkun,
  Chang, and Gupta}]{dhamala2021bold}
Jwala Dhamala, Tony Sun, Varun Kumar, Satyapriya Krishna, Yada Pruksachatkun,
  Kai-Wei Chang, and Rahul Gupta. 2021.
\newblock Bold: Dataset and metrics for measuring biases in open-ended language
  generation.
\newblock In \emph{Proceedings of the 2021 ACM conference on fairness,
  accountability, and transparency}, pages 862--872.

\bibitem[{Dwork et~al.(2012)Dwork, Hardt, Pitassi, Reingold, and
  Zemel}]{dwork2012fairness}
Cynthia Dwork, Moritz Hardt, Toniann Pitassi, Omer Reingold, and Richard Zemel.
  2012.
\newblock Fairness through awareness.
\newblock In \emph{Proceedings of the 3rd innovations in theoretical computer
  science conference}, pages 214--226.

\bibitem[{Fan et~al.(2018)Fan, Lewis, and Dauphin}]{fan2018hierarchical}
Angela Fan, Mike Lewis, and Yann Dauphin. 2018.
\newblock Hierarchical neural story generation.
\newblock \emph{arXiv preprint arXiv:1805.04833}.

\bibitem[{Gao et~al.(2021)Gao, Biderman, Black, Golding, Hoppe, Foster, Phang,
  He, Thite, Nabeshima, Presser, and Leahy}]{DBLP:Pile}
Leo Gao, Stella Biderman, Sid Black, Laurence Golding, Travis Hoppe, Charles
  Foster, Jason Phang, Horace He, Anish Thite, Noa Nabeshima, Shawn Presser,
  and Connor Leahy. 2021.
\newblock \href {http://arxiv.org/abs/2101.00027} {The pile: An 800gb dataset
  of diverse text for language modeling}.
\newblock \emph{CoRR}, abs/2101.00027.

\bibitem[{Geng and Liu(2023)}]{openlm2023openllama}
Xinyang Geng and Hao Liu. 2023.
\newblock \href {https://github.com/openlm-research/open_llama} {Openllama: An
  open reproduction of llama}.

\bibitem[{Guo and Caliskan(2021)}]{guo2021detecting}
Wei Guo and Aylin Caliskan. 2021.
\newblock Detecting emergent intersectional biases: Contextualized word
  embeddings contain a distribution of human-like biases.
\newblock In \emph{Proceedings of the 2021 AAAI/ACM Conference on AI, Ethics,
  and Society}, pages 122--133.

\bibitem[{Hardt et~al.(2016)Hardt, Price, and Srebro}]{hardt2016equality}
Moritz Hardt, Eric Price, and Nati Srebro. 2016.
\newblock Equality of opportunity in supervised learning.
\newblock \emph{Advances in neural information processing systems}, 29.

\bibitem[{Hoffmann et~al.(2022)Hoffmann, Borgeaud, Mensch, Buchatskaya, Cai,
  Rutherford, Casas, Hendricks, Welbl, Clark et~al.}]{hoffmann2022training}
Jordan Hoffmann, Sebastian Borgeaud, Arthur Mensch, Elena Buchatskaya, Trevor
  Cai, Eliza Rutherford, Diego de~Las Casas, Lisa~Anne Hendricks, Johannes
  Welbl, Aidan Clark, et~al. 2022.
\newblock Training compute-optimal large language models.
\newblock \emph{arXiv preprint arXiv:2203.15556}.

\bibitem[{Holtzman et~al.(2019)Holtzman, Buys, Du, Forbes, and
  Choi}]{holtzman2019curious}
Ari Holtzman, Jan Buys, Li~Du, Maxwell Forbes, and Yejin Choi. 2019.
\newblock The curious case of neural text degeneration.
\newblock \emph{arXiv preprint arXiv:1904.09751}.

\bibitem[{Hooker(2021)}]{hooker2021moving}
Sara Hooker. 2021.
\newblock Moving beyond “algorithmic bias is a data problem”.
\newblock \emph{Patterns}, 2(4).

\bibitem[{Huang et~al.(2019)Huang, Zhang, Jiang, Stanforth, Welbl, Rae, Maini,
  Yogatama, and Kohli}]{huang2019reducing}
Po-Sen Huang, Huan Zhang, Ray Jiang, Robert Stanforth, Johannes Welbl, Jack
  Rae, Vishal Maini, Dani Yogatama, and Pushmeet Kohli. 2019.
\newblock Reducing sentiment bias in language models via counterfactual
  evaluation.
\newblock \emph{arXiv preprint arXiv:1911.03064}.

\bibitem[{Jiang et~al.(2020)Jiang, Xu, Araki, and Neubig}]{jiang2020can}
Zhengbao Jiang, Frank~F Xu, Jun Araki, and Graham Neubig. 2020.
\newblock How can we know what language models know?
\newblock \emph{Transactions of the Association for Computational Linguistics},
  8:423--438.

\bibitem[{Klein et~al.(2017)Klein, Kim, Deng, Senellart, and
  Rush}]{klein2017opennmt}
Guillaume Klein, Yoon Kim, Yuntian Deng, Jean Senellart, and Alexander~M Rush.
  2017.
\newblock Opennmt: Open-source toolkit for neural machine translation.
\newblock \emph{arXiv preprint arXiv:1701.02810}.

\bibitem[{Liang et~al.(2022)Liang, Bommasani, Lee, Tsipras, Soylu, Yasunaga,
  Zhang, Narayanan, Wu, Kumar et~al.}]{liang2022holistic}
Percy Liang, Rishi Bommasani, Tony Lee, Dimitris Tsipras, Dilara Soylu,
  Michihiro Yasunaga, Yian Zhang, Deepak Narayanan, Yuhuai Wu, Ananya Kumar,
  et~al. 2022.
\newblock Holistic evaluation of language models.
\newblock \emph{arXiv preprint arXiv:2211.09110}.

\bibitem[{Lin et~al.(2021)Lin, Hilton, and Evans}]{lin2021truthfulqa}
Stephanie Lin, Jacob Hilton, and Owain Evans. 2021.
\newblock Truthfulqa: Measuring how models mimic human falsehoods.
\newblock \emph{arXiv preprint arXiv:2109.07958}.

\bibitem[{Loshchilov and Hutter(2017)}]{loshchilov2017decoupled}
Ilya Loshchilov and Frank Hutter. 2017.
\newblock Decoupled weight decay regularization.
\newblock \emph{arXiv preprint arXiv:1711.05101}.

\bibitem[{May et~al.(2019)May, Wang, Bordia, Bowman, and
  Rudinger}]{may2019measuring}
Chandler May, Alex Wang, Shikha Bordia, Samuel~R Bowman, and Rachel Rudinger.
  2019.
\newblock On measuring social biases in sentence encoders.
\newblock \emph{arXiv preprint arXiv:1903.10561}.

\bibitem[{Merity et~al.(2016)Merity, Xiong, Bradbury, and
  Socher}]{merity2016pointer}
Stephen Merity, Caiming Xiong, James Bradbury, and Richard Socher. 2016.
\newblock \href {http://arxiv.org/abs/1609.07843} {Pointer sentinel mixture
  models}.

\bibitem[{Nadeem et~al.(2020)Nadeem, Bethke, and Reddy}]{nadeem2020stereoset}
Moin Nadeem, Anna Bethke, and Siva Reddy. 2020.
\newblock Stereoset: Measuring stereotypical bias in pretrained language
  models.
\newblock \emph{arXiv preprint arXiv:2004.09456}.

\bibitem[{Nangia et~al.(2020)Nangia, Vania, Bhalerao, and
  Bowman}]{nangia2020crows}
Nikita Nangia, Clara Vania, Rasika Bhalerao, and Samuel~R Bowman. 2020.
\newblock Crows-pairs: A challenge dataset for measuring social biases in
  masked language models.
\newblock \emph{arXiv preprint arXiv:2010.00133}.

\bibitem[{Niu et~al.(2020)Niu, Yavuz, Zhou, Keskar, Wang, and
  Xiong}]{niu2020unsupervised}
Tong Niu, Semih Yavuz, Yingbo Zhou, Nitish~Shirish Keskar, Huan Wang, and
  Caiming Xiong. 2020.
\newblock Unsupervised paraphrasing with pretrained language models.
\newblock \emph{arXiv preprint arXiv:2010.12885}.

\bibitem[{Petroni et~al.(2019)Petroni, Rockt{\"a}schel, Lewis, Bakhtin, Wu,
  Miller, and Riedel}]{petroni2019language}
Fabio Petroni, Tim Rockt{\"a}schel, Patrick Lewis, Anton Bakhtin, Yuxiang Wu,
  Alexander~H Miller, and Sebastian Riedel. 2019.
\newblock Language models as knowledge bases?
\newblock \emph{arXiv preprint arXiv:1909.01066}.

\bibitem[{Radford et~al.(2019)Radford, Wu, Child, Luan, Amodei, Sutskever
  et~al.}]{radford2019language}
Alec Radford, Jeffrey Wu, Rewon Child, David Luan, Dario Amodei, Ilya
  Sutskever, et~al. 2019.
\newblock Language models are unsupervised multitask learners.
\newblock \emph{OpenAI blog}, 1(8):9.

\bibitem[{Rae et~al.(2021)Rae, Borgeaud, Cai, Millican, Hoffmann, Song,
  Aslanides, Henderson, Ring, Young et~al.}]{rae2021scaling}
Jack~W Rae, Sebastian Borgeaud, Trevor Cai, Katie Millican, Jordan Hoffmann,
  Francis Song, John Aslanides, Sarah Henderson, Roman Ring, Susannah Young,
  et~al. 2021.
\newblock Scaling language models: Methods, analysis \& insights from training
  gopher.
\newblock \emph{arXiv preprint arXiv:2112.11446}.

\bibitem[{Roche(2019)}]{roche2019articulating}
Gerald Roche. 2019.
\newblock Articulating language oppression: colonialism, coloniality and the
  erasure of tibet’s minority languages.
\newblock \emph{Patterns of prejudice}, 53(5):487--514.

\bibitem[{Rojas et~al.(2022)Rojas, Diamos, Kini, Kanter, Reddi, and
  Coleman}]{rojas2022dollar}
William A~Gaviria Rojas, Sudnya Diamos, Keertan~Ranjan Kini, David Kanter,
  Vijay~Janapa Reddi, and Cody Coleman. 2022.
\newblock The dollar street dataset: Images representing the geographic and
  socioeconomic diversity of the world.
\newblock In \emph{Thirty-sixth Conference on Neural Information Processing
  Systems Datasets and Benchmarks Track}.

\bibitem[{Romano et~al.(2006)Romano, Kouylekov, Szpektor, Dagan, and
  Lavelli}]{romano2006investigating}
Lorenza Romano, Milen~Ognianov Kouylekov, Idan Szpektor, Ido~Kalman Dagan, and
  Alberto Lavelli. 2006.
\newblock Investigating a generic paraphrase-based approach for relation
  extraction.
\newblock In \emph{11th Conference of the European Chapter of the Association
  for Computational Linguistics (EACL 2006)}, pages 409--416.

\bibitem[{Rudinger et~al.(2018)Rudinger, Naradowsky, Leonard, and
  Van~Durme}]{rudinger2018gender}
Rachel Rudinger, Jason Naradowsky, Brian Leonard, and Benjamin Van~Durme. 2018.
\newblock Gender bias in coreference resolution.
\newblock \emph{arXiv preprint arXiv:1804.09301}.

\bibitem[{Scao et~al.(2022)Scao, Fan, Akiki, Pavlick, Ili{\'c}, Hesslow,
  Castagn{\'e}, Luccioni, Yvon, Gall{\'e} et~al.}]{scao2022bloom}
Teven~Le Scao, Angela Fan, Christopher Akiki, Ellie Pavlick, Suzana Ili{\'c},
  Daniel Hesslow, Roman Castagn{\'e}, Alexandra~Sasha Luccioni, Fran{\c{c}}ois
  Yvon, Matthias Gall{\'e}, et~al. 2022.
\newblock Bloom: A 176b-parameter open-access multilingual language model.
\newblock \emph{arXiv preprint arXiv:2211.05100}.

\bibitem[{Schw{\"o}bel(2022)}]{schwobel2022learned}
Pola Schw{\"o}bel. 2022.
\newblock Learned data augmentation for bias correction.
\newblock \emph{PhD Thesis}.

\bibitem[{Schw{\"o}bel and Remmers(2022)}]{schwobel2022long}
Pola Schw{\"o}bel and Peter Remmers. 2022.
\newblock The long arc of fairness: Formalisations and ethical discourse.
\newblock In \emph{2022 ACM Conference on Fairness, Accountability, and
  Transparency}, pages 2179--2188.

\bibitem[{Sheng et~al.(2021)Sheng, Chang, Natarajan, and
  Peng}]{sheng2021societal}
Emily Sheng, Kai-Wei Chang, Premkumar Natarajan, and Nanyun Peng. 2021.
\newblock Societal biases in language generation: Progress and challenges.
\newblock \emph{arXiv preprint arXiv:2105.04054}.

\bibitem[{Tan and Celis(2019)}]{tan2019assessing}
Yi~Chern Tan and L~Elisa Celis. 2019.
\newblock Assessing social and intersectional biases in contextualized word
  representations.
\newblock \emph{Advances in neural information processing systems}, 32.

\bibitem[{Touvron et~al.(2023)Touvron, Lavril, Izacard, Martinet, Lachaux,
  Lacroix, Rozi{\`e}re, Goyal, Hambro, Azhar et~al.}]{touvron2023llama}
Hugo Touvron, Thibaut Lavril, Gautier Izacard, Xavier Martinet, Marie-Anne
  Lachaux, Timoth{\'e}e Lacroix, Baptiste Rozi{\`e}re, Naman Goyal, Eric
  Hambro, Faisal Azhar, et~al. 2023.
\newblock Llama: Open and efficient foundation language models.
\newblock \emph{arXiv preprint arXiv:2302.13971}.

\bibitem[{{US Equal Employment Opportunity Commission}(1979)}]{us1979questions}
{US Equal Employment Opportunity Commission}. 1979.
\newblock Questions and answers to clarify and provide a common interpretation
  of the uniform guidelines on employee selection procedures.
\newblock \emph{Federal Register}, 40(43).

\bibitem[{Zaken et~al.(2021)Zaken, Ravfogel, and Goldberg}]{zaken2021bitfit}
Elad~Ben Zaken, Shauli Ravfogel, and Yoav Goldberg. 2021.
\newblock Bitfit: Simple parameter-efficient fine-tuning for transformer-based
  masked language-models.
\newblock \emph{arXiv preprint arXiv:2106.10199}.

\bibitem[{Zhou et~al.(2022)Zhou, Ethayarajh, and Jurafsky}]{zhou2022richer}
Kaitlyn Zhou, Kawin Ethayarajh, and Dan Jurafsky. 2022.
\newblock Richer countries and richer representations.
\newblock In \emph{Findings of the Association for Computational Linguistics:
  ACL 2022}, pages 2074--2085.

\end{thebibliography}
\bibliographystyle{acl_natbib}

\clearpage

\appendix
\section{Choosing $r$ -- additional Models}
\label{sec:appendixA}

\begin{figure*}
     \centering
     \begin{subfigure}[b]{0.32\textwidth}
         \centering
         \includegraphics[width=\textwidth]{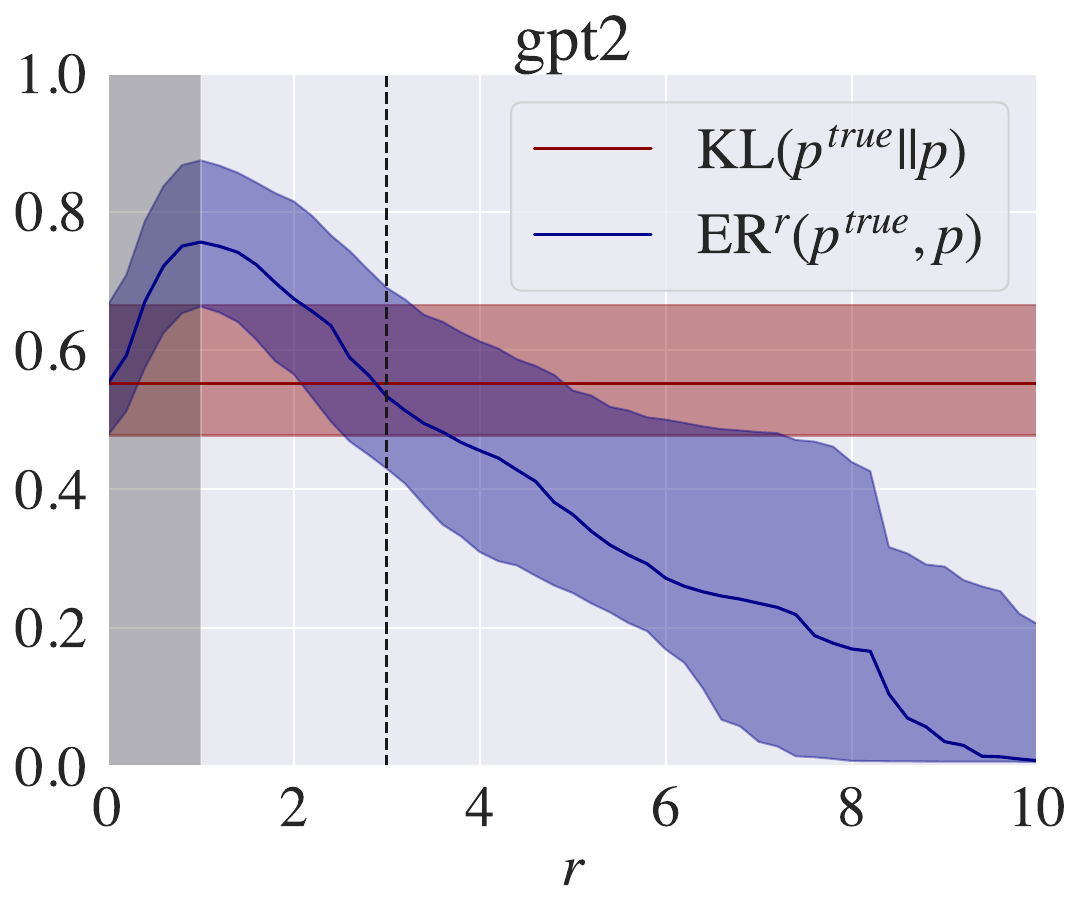}
         %\caption{GPT2}
     \end{subfigure}
     \hfill
     \begin{subfigure}[b]{0.32\textwidth}
         \centering
         \includegraphics[width=\textwidth]{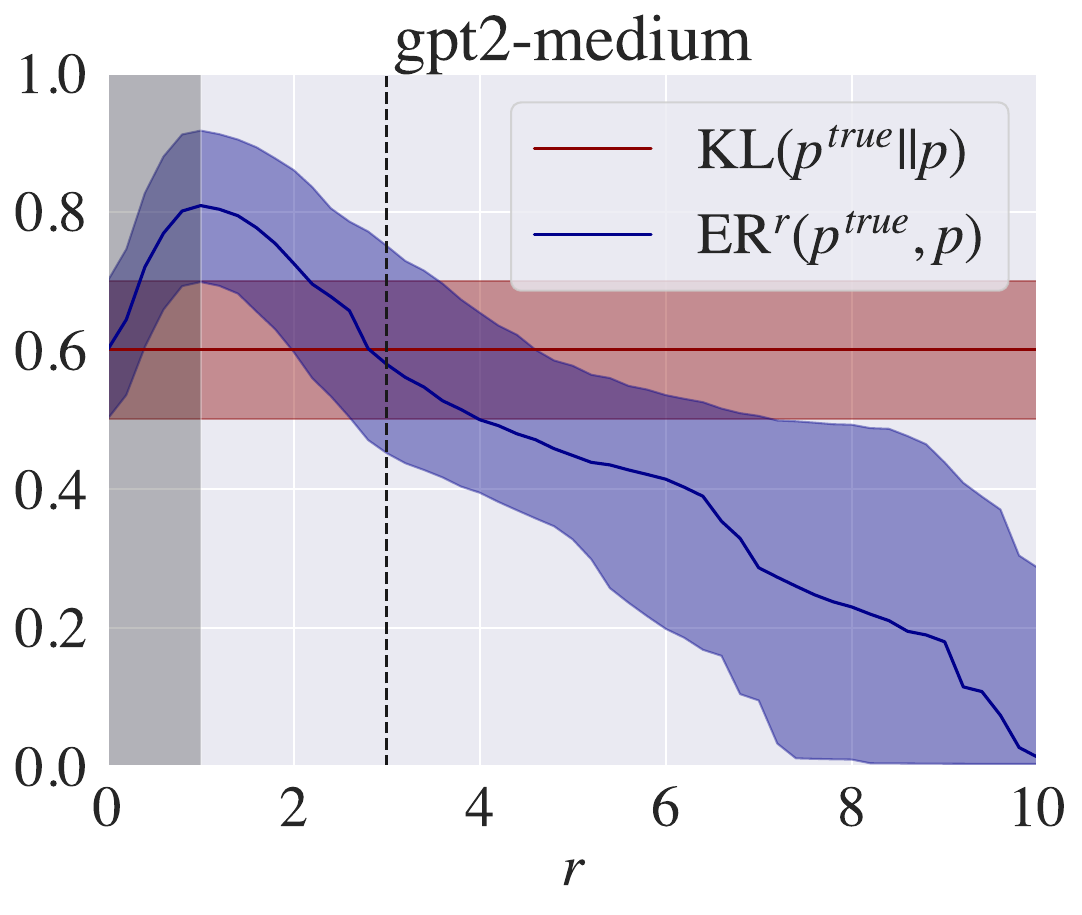}
         %\caption{GPT2-medium}
     \end{subfigure}
     \hfill
     \begin{subfigure}[b]{0.32\textwidth}
         \centering
         \includegraphics[width=\textwidth]{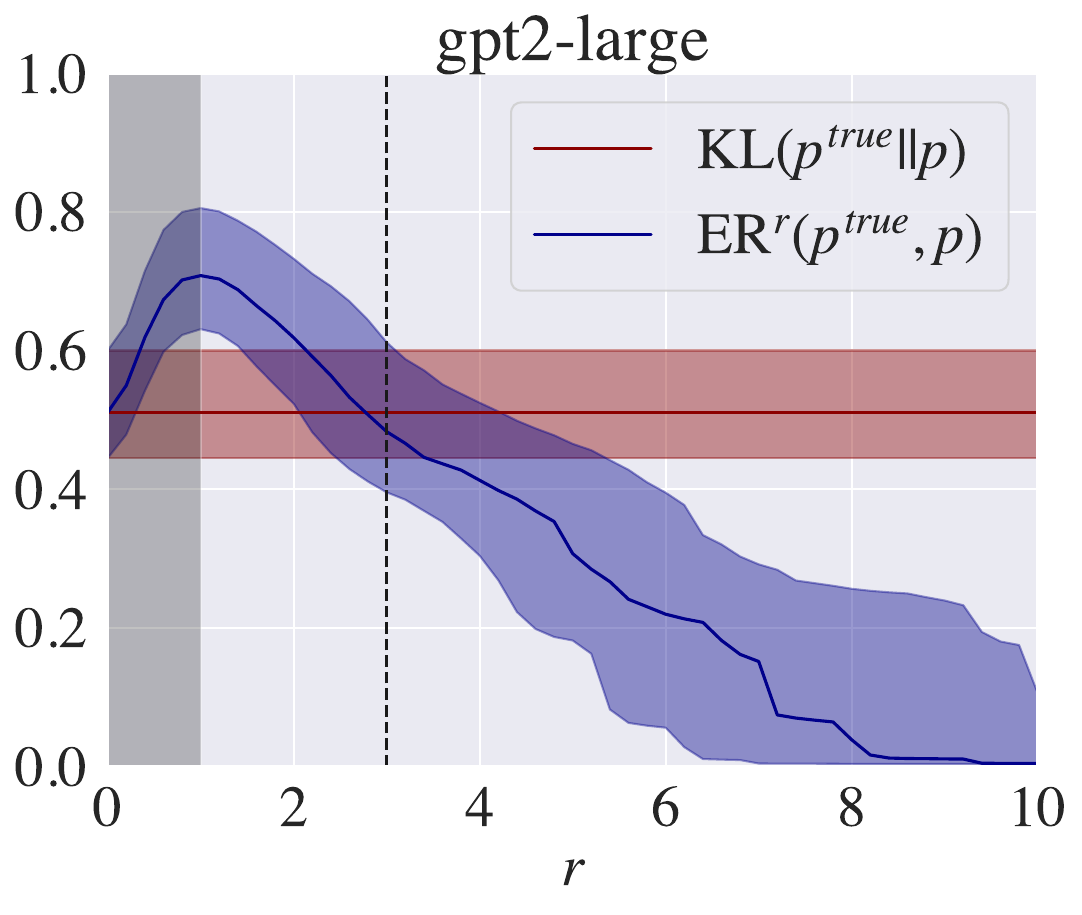}
         %\caption{GPT2-large}
     \end{subfigure}
     %linebreak 

     \begin{subfigure}[b]{0.32\textwidth}
         \centering
         \includegraphics[width=\textwidth]{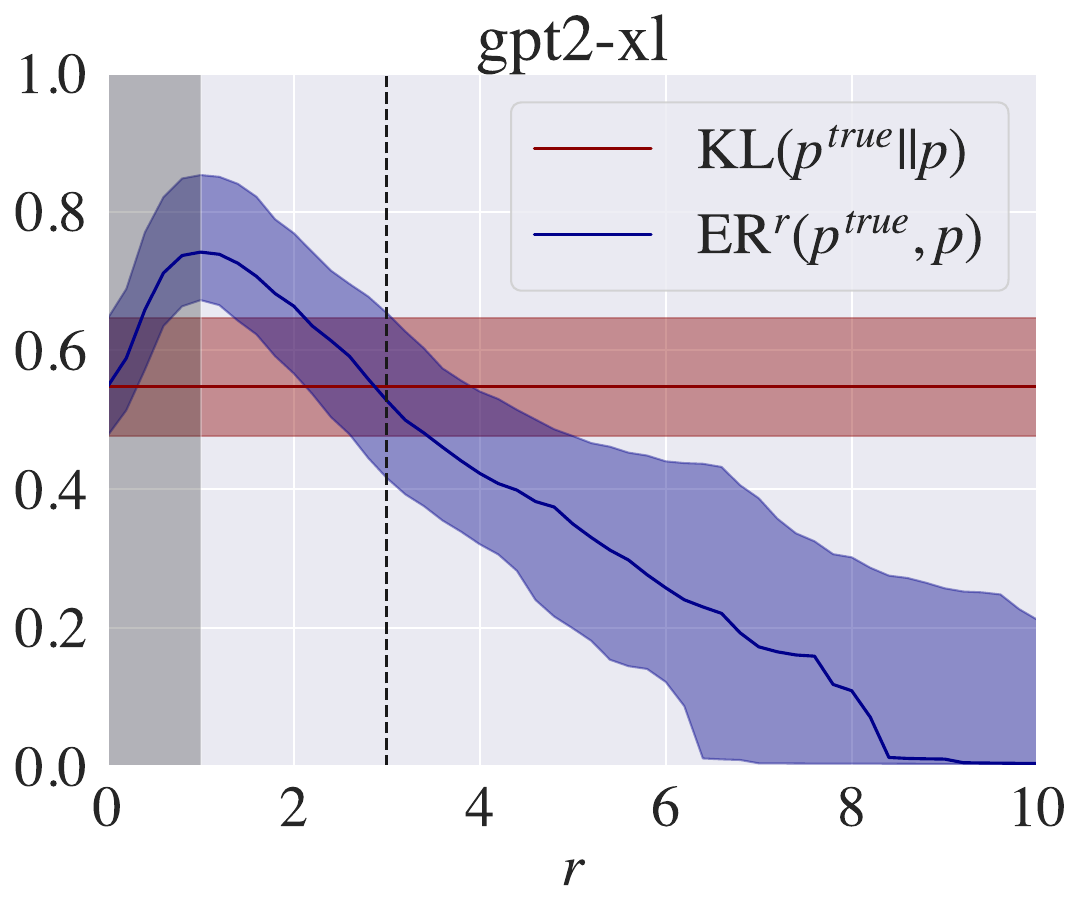}
         %\caption{GPT2-xl}
     \end{subfigure}
    \hfill
    \begin{subfigure}[b]{0.32\textwidth}
         \centering
         \includegraphics[width=\textwidth]{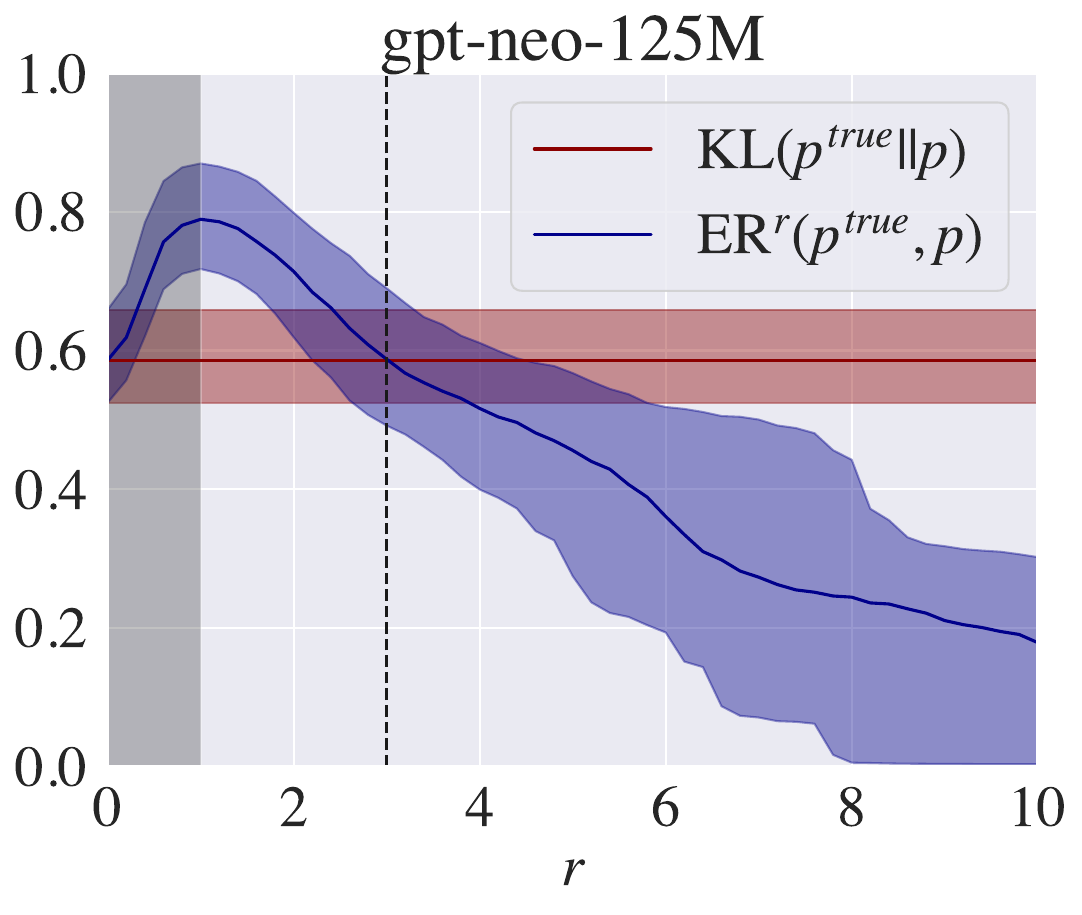}
         %\caption{GPT-Neo-125M}
     \end{subfigure}
    \hfill
     \begin{subfigure}[b]{0.32\textwidth}
         \centering
         \includegraphics[width=\textwidth]{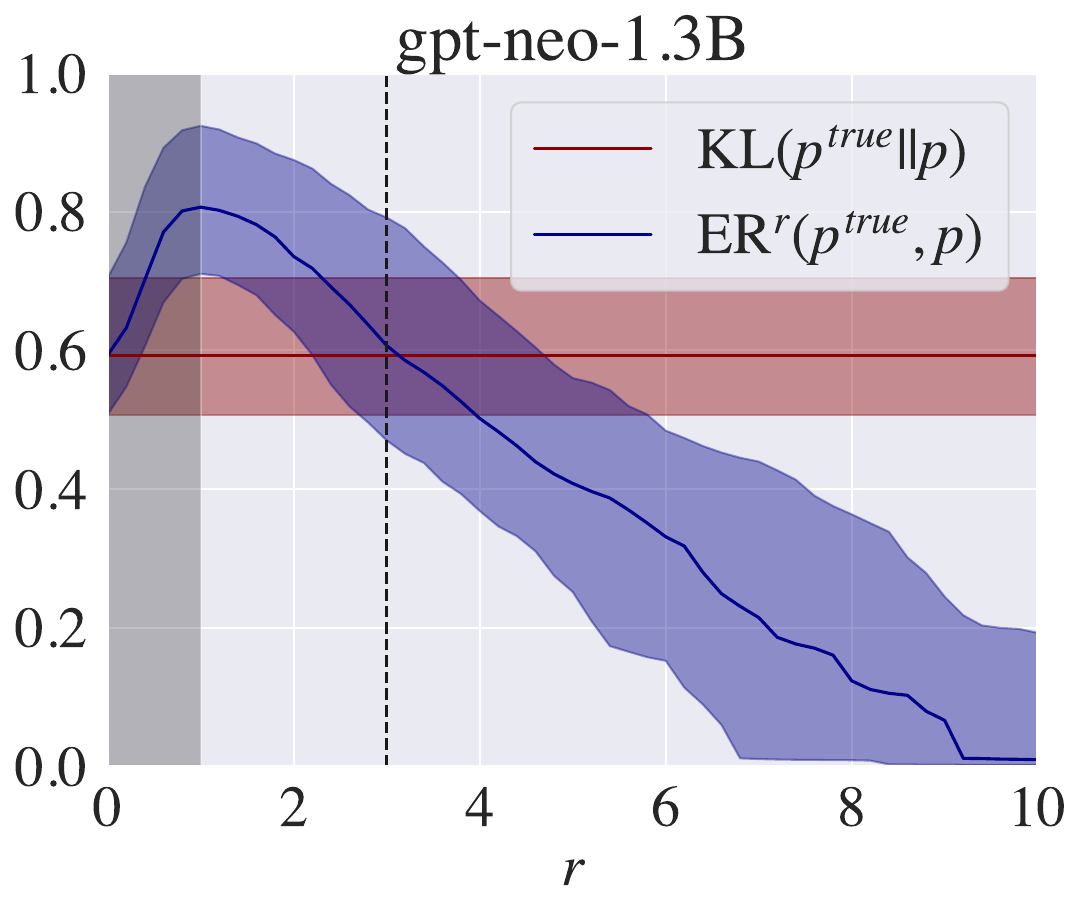}
         %\caption{GPT-Neo-1.3B}
     \end{subfigure}
    % linebreak
     
     \begin{subfigure}[b]{0.32\textwidth}
         \centering
         \includegraphics[width=\textwidth]{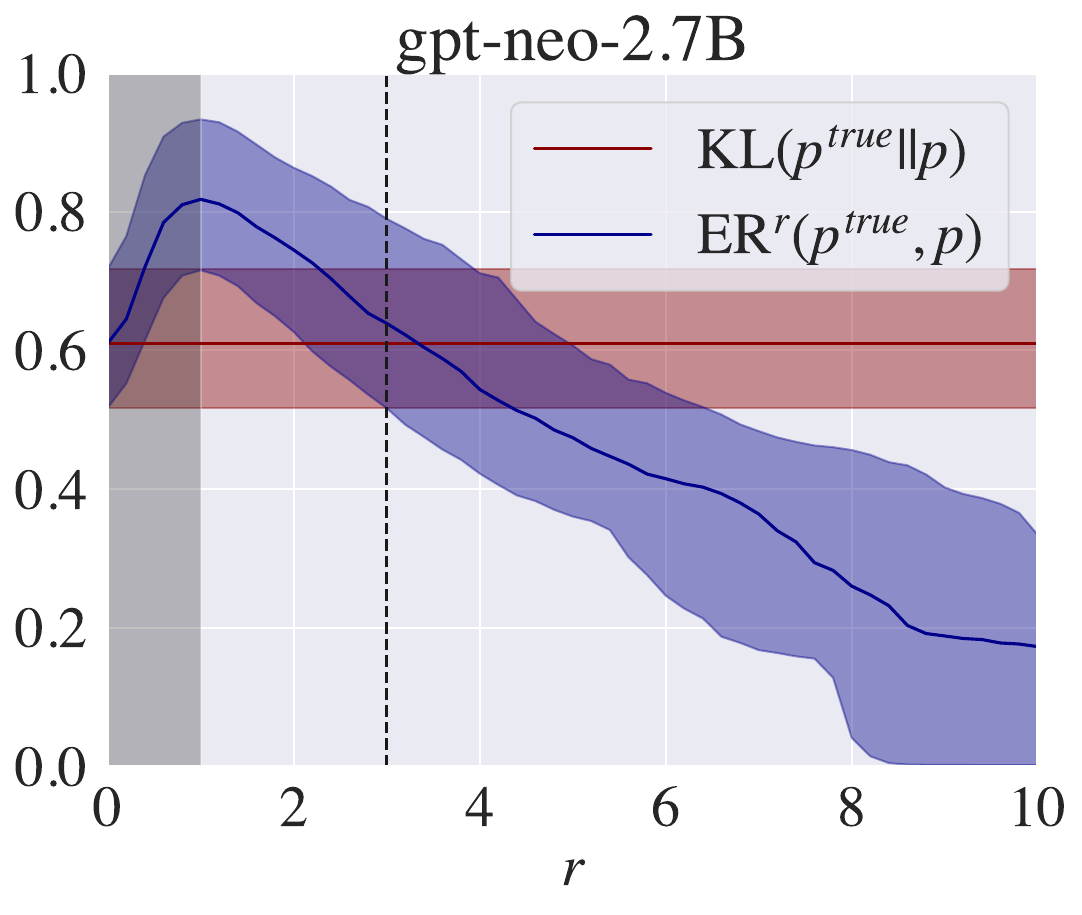}
         %\caption{GPT-Neo-2.7B}
     \end{subfigure}
    \hfill
     \begin{subfigure}[b]{0.32\textwidth}
         \centering
         \includegraphics[width=\textwidth]{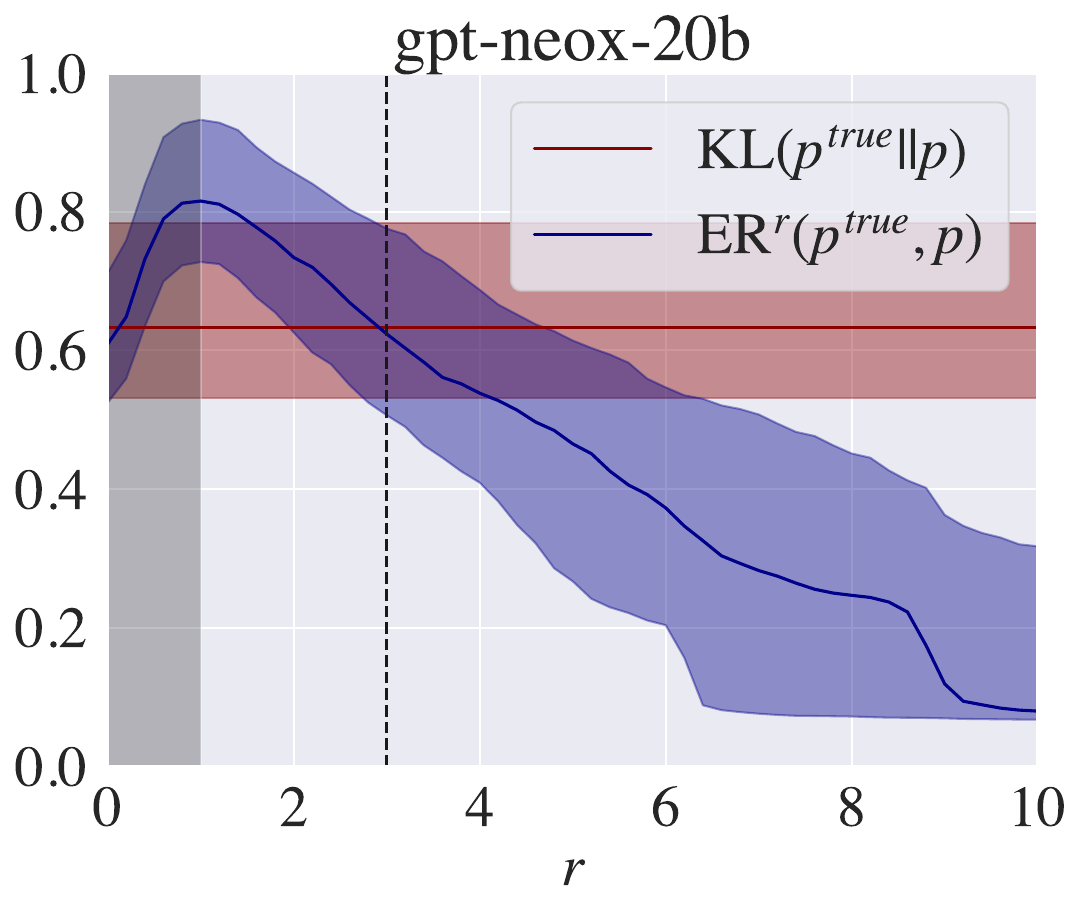}
         %\caption{GPT-NeoX-20B}
     \end{subfigure}
     \hfill
     \begin{subfigure}[b]{0.32\textwidth}
         \centering
         \includegraphics[width=\textwidth]{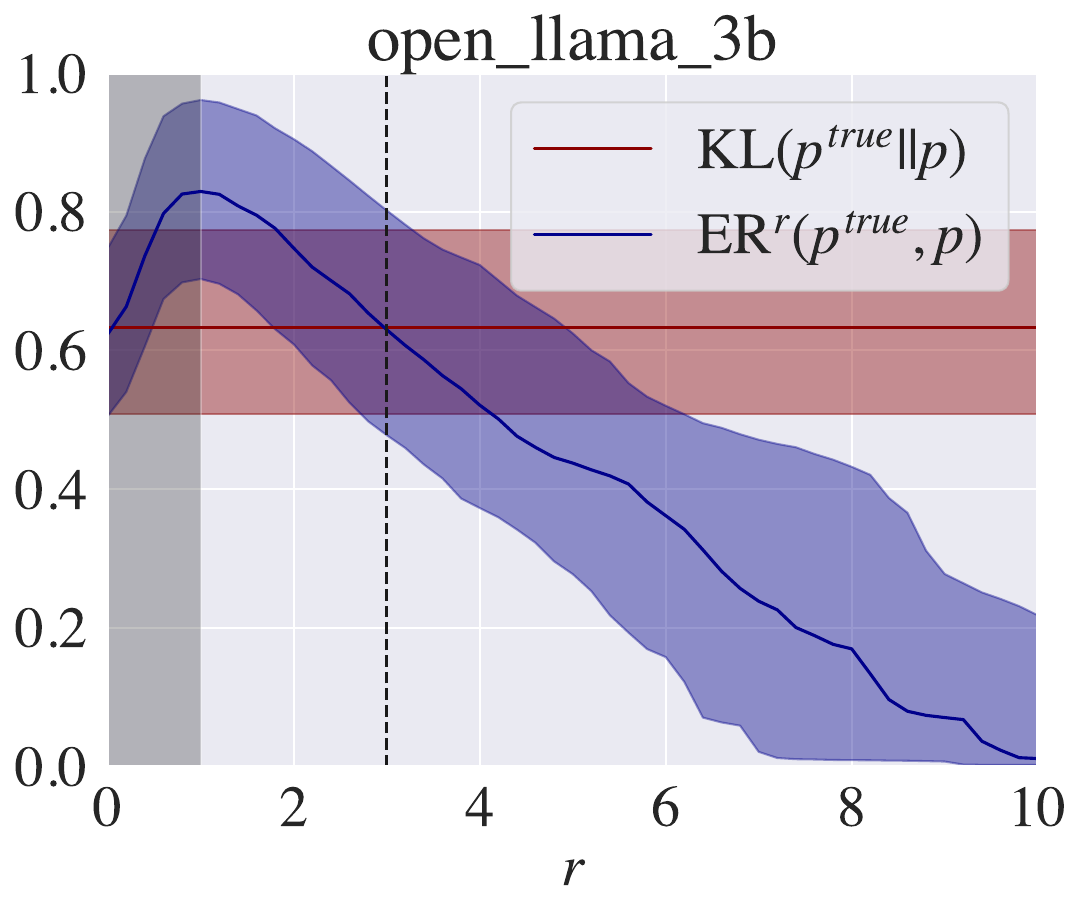}
         %\caption{OpenLLaMA 3B}
     \end{subfigure}
        \caption{\textbf{The relationship between ER$^r$ and KL-divergence is similar for all models.} We compare ER$^r$ (blue) for different $r$ to the KL-divergence (red), repeating the experiment from \S\ref{sec:properties} for all models. The median is plotted as a solid line, $25^{th} - 75^{th}$ percentiles as shaded areas (over different versions of the ``I live in'' -prompt, see \S\ref{sec:rephrasing}). We pick $r=3$, the integer value for which KL and ER$^r$ are the most similar for all models (dashed line, black).}
        \label{fig:appendix_ebr_kl}
\end{figure*}

Section \ref{sec:properties} compares ER$^r$ for different values of $r$ to the KL-divergence. We pick $r=3$ in this experiment such that $\text{ER}(\ptrue, p)^r \approx \text{KL}(\ptrue || p)$. Figure \ref{fig:appendix_ebr_kl} contains the same experiment for all models under consideration. The optimal choice according to this heuristic is $r=3$ for all of them.

\section{Details on Prompt Rephrasing} \label{sec:appendixB}

As described in \S\ref{sec:rephrasing}, we formulate different versions of a seed prompt $c=$``I am from \_\_\_.'' in order to produce a set of diverse prompt wordings $\mathcal{D}$. 
We do this in two steps. First, we prompt ChatGPT\footnote{Accessed via https://chat.openai.com/.} to rephrase the seed prompt ``I am from''. We manually remove irrelevant prompts, e.g.. those containing country adjectives rather than nouns. This first step extends the seed prompt into to the following $16$ base formulations.

\subsection{List of prompts}
\begin{enumerate}
\item I am from \_\_\_.
\item I live in \_\_\_.
\item I hail from \_\_\_.
\item My homeland is \_\_\_.
\item I come from \_\_\_.
\item I was born and raised in \_\_\_.
\item I am a citizen of \_\_\_.
\item I originate from \_\_\_.
\item My roots are in \_\_\_.
\item I grew up in \_\_\_.
\item I was brought up in \_\_\_.
\item I was raised in \_\_\_.
\item I was born in \_\_\_.
\item My place of origin is \_\_\_.
\item I reside in \_\_\_.
\item My home country is \_\_\_.
\end{enumerate}

Secondly, we apply template-based rephrasing similar to \citet{romano2006investigating}. We replace the subject of the sentence from "I"  to alternative pronouns \{"She", "He", "We", "They", "You"\} -- adjusting verb forms accordingly -- as well as combinations of \{"My", "Her", "His", "Our", "Their"\} $\times$ \{ "uncle",
        "aunt",
        "brother",
        "sister",
        "niece",
        "nephew",
        "mother",
        "father",
        "mom",
        "daughter",
        "son",
        "cousin",
        "friend",
        "relative"\}.
This second step extends the $16$ base formulations into our final dataset $\mathcal{D}$ containing $955$ prompts.

\section{Variability across Prompts} \label{sec:appendixC}
Recall from \eqref{eq:uniform_agg} that 
\begin{align*}
    p_i^{\text{agg\_uni}} = p(x_i| \mathcal{M}) &= \mathds{E}_{p(c | \mathcal{M})} \left [ p(x_i |c) \right ]
\end{align*}
under uniform $p(c | \mathcal{M})$. By Jensen's inequality,
\begin{align*}
    \text{ER}^r(p^\text{true}, & \pagguni) = \\
    & \sum_{i \in \Sr} \ptrue_i \log \left( \frac{\ptrue_i}{\mathds{E}_{p(c | \mathcal{M})} \left [ p(x_i |c) \right ]} \right) \\
    & \leq \mathds{E}_{p(c | \mathcal{M})} \left[ \sum_{i \in \Sr} \ptrue_i \log \left( \frac{\ptrue_i}{p(x_i |c)} \right) \right] \\
    &= \frac{1}{C} \sum_{c \in \mathcal{D}} \text{ER}^r(\ptrue_i, p( x_i | c)).
\end{align*}
Thus, erasure under the aggregate distribution $\text{ER}^r(p^\text{true}, \pagguni)$ is  a lower bound to the average erasure. 

\section{Alternative Mitigation Strategies} \label{sec:appendixD}
\textbf{Finetuning for other values of $r$}: In \S\ref{sec:exp4} we mitigate erasure by finetuning, employing ER$^3$ as a loss function (Figure \ref{fig:mitigation_exp}). This choice corresponds to a minimal intervention where we only modify the distributions for affected countries at a rate above $r=3$; we do not address any underprediction by a smaller degree. 

\begin{figure*}
\centering
\includegraphics[width=\textwidth]{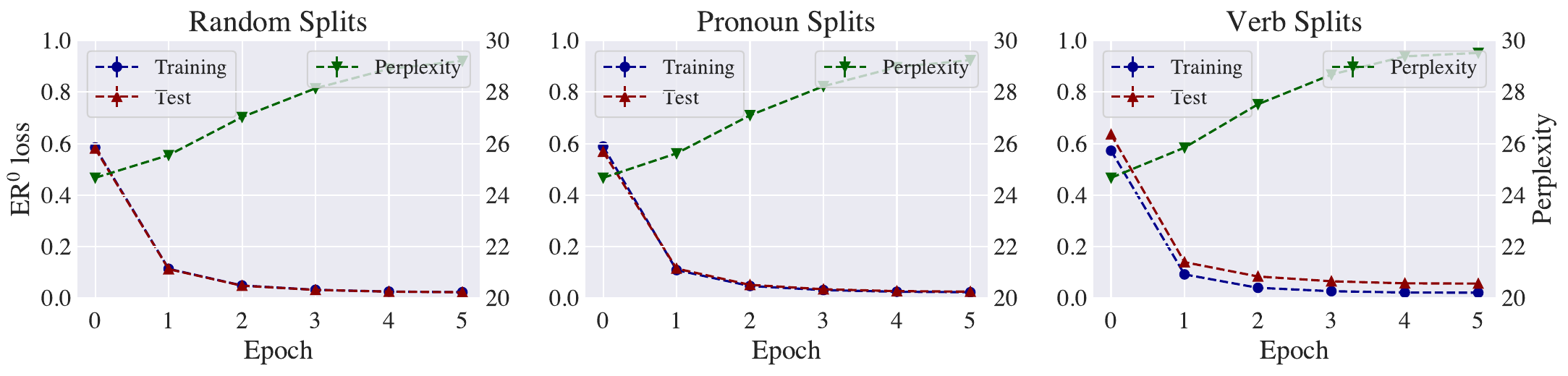} 
\caption{\textbf{Finetuning at $r=0$ also mitigates erasure, though at a higher perplexity cost.} Like in Figure \ref{fig:mitigation_exp} we plot average ER$^r$ on training (blue) and test (red) set prompts during $5$ epochs of finetuning of the GPT2-small model. Error bars indicate minima/maxima over $5$ folds. Note the different y-axis scale compared to Figure \ref{fig:appendix_mitigation_exp}.
} \label{fig:appendix_mitigation_exp}
\end{figure*}

Finetuning with $r=0$ (Fig.~\ref{fig:appendix_mitigation_exp}) is a stronger intervention, matching the distributions for \textit{all} countries (since EB$^0(\ptrue, p) = \text{KL}(\ptrue||p)$, see \S \ref{sec:properties}). As before, we can match the full distributions and achieve EB$^0(\ptrue, p) \approx 0$ after only $5$ epochs of finetuning. However, due to the more drastic intervention into the model distribution $\ptrue$ the drop in language modelling performance is larger. Perplexity increases by almost $20$\% compared to $5$\% in Figure \ref{fig:mitigation_exp}. Note the different y-axis scales between Figures \ref{fig:mitigation_exp} and \ref{fig:appendix_mitigation_exp}.

\begin{table*}
    \centering
    \begin{tabular}{|c||l|c|c|c|} \hline 
          &No mitigation&  ER$^3$&  ER$^0$& $\tau$\\ \hline \hline
         Training loss &0.5722 $\pm$ 0.0160&  \textbf{0.0054} $\pm$ \textbf{0.0003} &  0.0211 $\pm$ 0.0003&  0.5641\\ \hline 
         Test loss &0.6065 $\pm$ 0.0543&  \textbf{0.0068} $\pm$ \textbf{0.0014} &   0.0566 $\pm$ 0.0020& -- \\ \hline 
         Perplexity &24.6716 $\pm$ 0&  25.8694 $\pm$ 0.0913&  29.5201 $\pm$ 0.0297& \textbf{24.5357}\\ \hline
    \end{tabular}
    \caption{\textbf{Summary of mitigation experiments.} Finetuning with ER$^3$, ER$^0$ and mitigation via optimising $\tau$. We report the training and test loss as well as perplexity after the 5$^{\text{th}}$ finetuning epoch. Best values bolded.}
    \label{tab:mitigation}
\end{table*}

\textbf{Mitigation via Temperature Softmax}:
A simple way to modify the model distribution $p$ % and hence erasure ER$^r(\ptrue, p)$ 
is via the softmax temperature parameter $\tau$ of the model. We have used $\tau = 1$ in all previous experiments. Here, we experiment with modifying $\tau$ to mitigate ER$^r(\ptrue, p)$ such that

\begin{align}
\text{ER}^r = \min_{\tau} \text{ER}^r(\ptrue, p_{\tau}). 
\end{align}

\begin{figure}
\centering
\includegraphics[width=\linewidth]{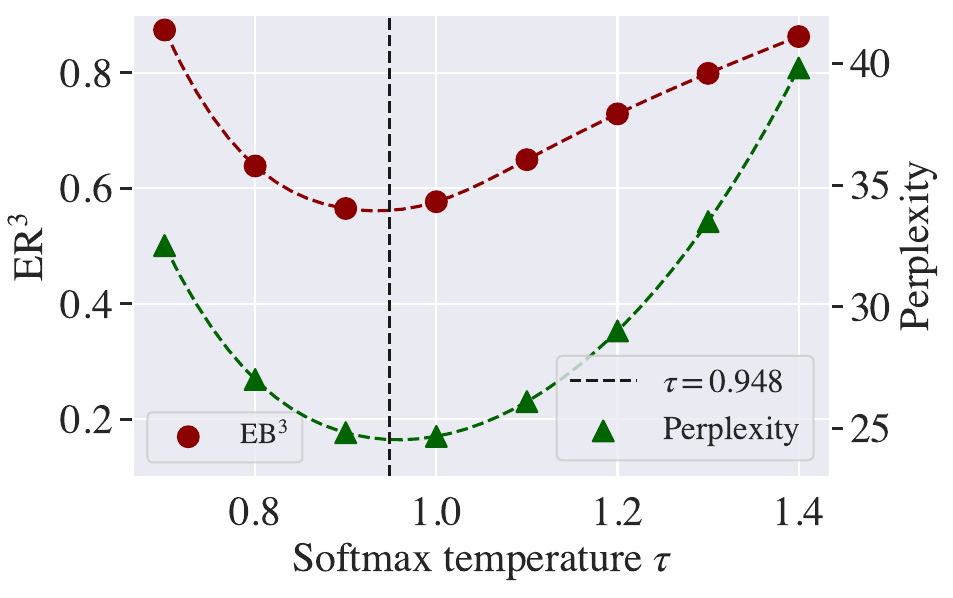} 
\caption{\textbf{Mitigating $\text{ER}^r$ using the temperature parameter $\tau$ is less successful than full finetuning.} $\text{ER}^r$ and perplexity are plotted as a function of $\tau$. The optimal value (minimising $\text{ER}^r$ w.r.t.~$\tau$) is $0.948$, dashed line.}
 \label{fig:temp_mitigation_exp}
\end{figure}

Figure \ref{fig:temp_mitigation_exp} shows $\text{ER}^r$ and perplexity as a function of $\tau$. The optimal value (minimising $\text{ER}^r$ w.r.t.~$\tau$) is $0.948$, dashed line. This mitigation method is compared to fine-tuning of the neural network parameters from earlier experiments in Table \ref{tab:mitigation}. The two middle columns correspond to the finetuning results from Figure \ref{fig:mitigation_exp} and Figure \ref{fig:appendix_mitigation_exp}, the rightmost column contains the results for varying temperature parameter $\tau$. 

Perhaps unsurprisingly, mitigation attempts with a single parameter $\tau$ are much less successful than using full finetuning (small drop in $\text{ER}^r$ only, see first row of Table \ref{tab:mitigation}). Perplexity, however, improves slightly over the original model. 

\end{document}